\documentclass{article}

\usepackage{microtype}
\usepackage{graphicx}
\usepackage{subfigure}
\usepackage{booktabs} 
\usepackage[utf8]{inputenc} 
\usepackage{xcolor}
\usepackage{ragged2e}
\usepackage[ruled,vlined]{algorithm2e}
\usepackage{amsmath}
\usepackage{mathtools}
\usepackage{mdframed}
\usepackage{soul}

\usepackage{tcolorbox}
\usepackage{tabularx}
\usepackage{amssymb}
\usepackage{pifont}
\colorlet{punct}{red!60!black}
\definecolor{background}{HTML}{EEEEEE}
\definecolor{delim}{RGB}{20,105,176}
\colorlet{numb}{magenta!60!black}

\usepackage{hyperref}

\sethlcolor{lightgray}
\definecolor{lightgreen}{rgb}{0.56, 0.93, 0.56}
\definecolor{deepblue}{HTML}{003D79}

\newcommand{\hlred}[1]{\sethlcolor{pink}\hl{#1}}
\newcommand{\hlgreen}[1]{\sethlcolor{lightgreen}\hl{#1}}
\newcommand{\hlyellow}[1]{\sethlcolor{yellow}\hl{#1}}
\usepackage[accepted]{icml2025}

\usepackage{amsmath}
\usepackage{amssymb}
\usepackage{mathtools}
\usepackage{amsthm}

\usepackage[capitalize,noabbrev]{cleveref}

\theoremstyle{plain}

\theoremstyle{definition}

\theoremstyle{remark}

\usepackage[textsize=tiny]{todonotes}

\icmltitlerunning{Submission and Formatting Instructions for ICML 2025}

\begin{document}

\twocolumn[
\icmltitle{General Modular Harness for LLM Agents in Multi-Turn Gaming Environments}



\icmlsetsymbol{equal}{*}

\begin{icmlauthorlist}
\icmlauthor{Yuxuan Zhang}{equal,hdsi}
\icmlauthor{Haoyang Yu}{equal,hdsi}
\icmlauthor{Lanxiang Hu}{hdsi}
\icmlauthor{Haojian Jin}{hdsi}
\icmlauthor{Hao Zhang}{hdsi}
\end{icmlauthorlist}

\icmlaffiliation{hdsi}{Halıcıoğlu Data Science Institute (HDSI), University of California San Diego, La Jolla, CA, USA}

\icmlcorrespondingauthor{Yuxuan Zhang}{yuz165@ucsd.edu}
\icmlcorrespondingauthor{Haoyang Yu}{hay034@ucsd.edu}

\icmlkeywords{Machine Learning, ICML}

\vskip 0.3in
]



\printAffiliationsAndNotice{\icmlEqualContribution} 

\begin{abstract}

We introduce a modular harness design for LLM agents that composes of perception, memory, and reasoning components, enabling a single LLM/VLM backbone to tackle a wide spectrum of multi-turn gaming environments without domain-specific engineering. Using classic and modern game suites as low-barrier, high-diversity testbeds, our framework provides a unified workflow for analyzing how each module affects performance across dynamic interactive settings. Extensive experiments demonstrate that the harness lifts gameplay performance consistently over un-harnessed baselines and reveals distinct contribution patterns—for example, memory dominates in long-horizon puzzles while perception is critical in vision-noisy arcades. 
These findings highlight the effectiveness of our modular harness design in advancing general-purpose agent, given the familiarity and ubiquity of games in everyday human experience. 
\end{abstract}
\section{Introduction}

LLM/VLM agents have shown promising results on multi-turn tasks involving web and desktop automation, as demonstrated by benchmarks like \textsc{WebArena}, \textsc{Mind2Web}, and \textsc{OSWorld}~\citep{webarena2024,deng2023mind2web,osworld2024}.
In general, their successes rely on domain-specific, hand-engineered workflows: systems such as \textsc{AutoGPT}, \textsc{LangGraph}, \textsc{Graph-of-Thoughts} chain together dozens of tool calls, UI heuristics, and prompt templates—each tailored to a specific environment~\citep{yang2023autogpt,langgraph2024,besta2024got}. %

Despite of its efficacy, domain‑customized workflow poses a few challenges. First, this design limits generalization: workflows hardcoded for office macros in \textsc{OSWorld} or site‑specific interactions in \textsc{WebArena} fail to transfer across tasks~\citep{osworld2024,shen2024scribeagent}.
Second, these agent workflows are structurally intricate, with perception, memory, and reasoning components deeply obscured—making it nearly impossible to isolate each module’s contribution to overall performance~\citep{wang2024awm,tan2024cradle}. %
Third, these tasks often require expertise knowledge like computer-use skills (e.g., managing Excel, PowerPoint, or system utilities), which raises the bar for both agent competence and human evaluation. %
These limitations expose a fundamental gap: the field lacks a principled, systematic method for studying modular agent design in general-purpose, multi-turn environments. As natively trained agentic models grow more capable with less need for hand-engineered behaviors~\citep{wang2025ragen, zhou2025sweet, NEURIPS2024rl4vlm, abdulhai2023lmrl}, such a framework becomes increasingly essential toward general-purpose agents. Interactive gaming environments offer a compelling testbed for studying this problem, providing (1) well-defined reward signals for evaluating harness effectiveness, (2) a diverse range of task settings, and (3) challenging gaming objectives.

Inspired by the open-ended methodology of \textsc{Voyager} in Minecraft~\citep{wang2023voyager}, we repurpose a wide range of classic and modern game suites as a unified laboratory for studying modular agent design.
First, these games offer a uniform control interface across diverse multi-turn interactive environments built on the standardized \textsc{Gymnasium} API~\citep{towers2024gymnasium} , enabling reproducible evaluation with minimal integration overhead. In addition, we integrate \textsc{Stable Retro}, which presents classic console ROMs through the same Gym interface, allowing our pipeline to extend seamlessly from grid-based titles like \textsc{Tetris} to richly animated franchises such as \textsc{Pokémon}.%

\begin{figure*}
    \centering
    \includegraphics[width=\textwidth]{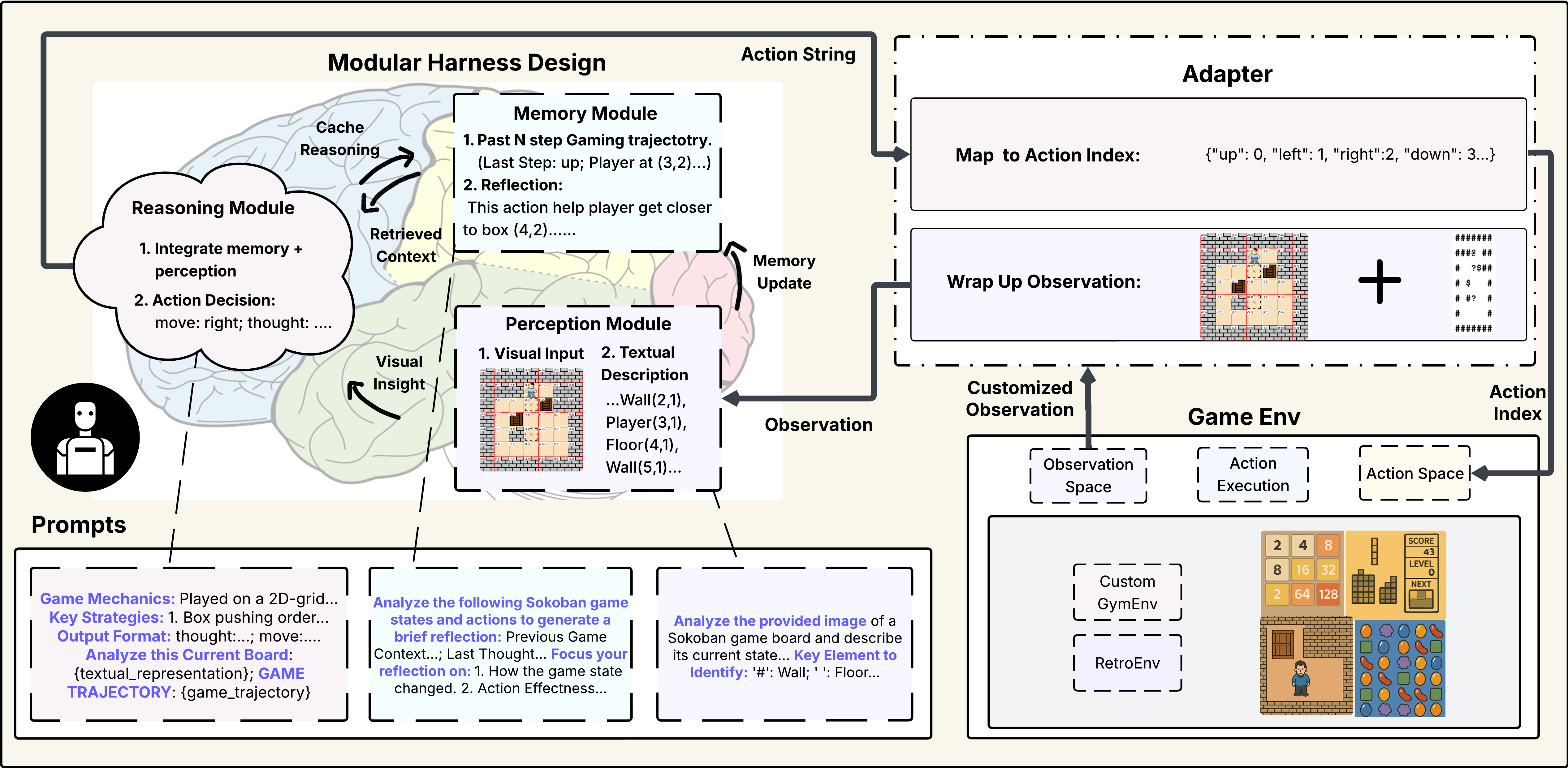}
    \caption{ 
    Our agent uses modular harnesses—such as perception, memory, and reasoning modules—to systematically extend a model’s game-playing capabilities, allowing the model to engage with a simulated game environment through iterative interaction loops.
    }
    \label{fig:game_and_overview}
\end{figure*}  

Second, games serve as clean abstractions of many different multi-turn interactive tasks: accommodating different mechanics, objectives, and dynamics within a shared multi-turn interaction structure makes them a universal challenge for general-purpose agents. %
Third, unlike real-world domains that demand specialized knowledge (e.g., office software or system administration), games are designed for rapid human learnability—their rules are simple, intuitive, and accessible to non-experts. %
This simplicity ensures that evaluations primarily reflect an agent’s core cognitive abilities—perception, memory, and reasoning—rather than reliance on domain-specific hacks. %

To ground our study, we select four widely recognised titles—\textsc{Sokoban}, \textsc{Candy~Crush}, \textsc{2048}, \textsc{Tetris}—that cover a broad spectrum of multi-turn interaction patterns. %
All four are computationally challenging: \textsc{Sokoban} and \textsc{Candy Crush} are NP-hard~\citep{SokobanNPhard1999,CandyCrushNPhard2014}, \textsc{2048} and \textsc{Tetris} are NP-complete~\citep{2048Hard2018,Demaine2003TetrisHard} %
Together they challenge models' spatial reasoning, long-horizon planning across diverse multi-turn environments %
 
Guided by Newell’s Unified Theories of Cognition—which posits perception, memory, and reasoning as core, interlocking faculties~\citep{NewellUTC1990}—we build a three-module harness atop a single backbone model. %
\textbf{Perception} helps agents process both low-resolution grid environments and visually complex, high-detail images (e.g., \textsc{Sokoban} and \textsc{Tetris}).%
\textbf{Memory} stores recent trajectories and synthesises \emph{self-reflection} signals, akin to the Reflexion paradigm, enabling the agent to critique past moves and adjust future plans~\citep{shinn2023reflexion}. 
\textbf{Reasoning} integrates perceptual embeddings and memory traces within the backbone LLM/VLM to produce sequential decisions. %
By toggling these modules on and off in controlled ablations, we can quantify each component’s marginal contribution to model performances across all four games, yielding a systematic, cross-domain analysis of modular agent design. %

In our empirical findings, across all four games, the full modular harness significantly improves mean win rate over the unharnessed LLM baseline (paired t-test, $p < 0.05$). %

Ablation studies reveal complementary effects: perception-only configurations benefit games like \textsc{Sokoban} and \textsc{Tetris}, which involve large grid-based boards with complex spatial layouts. In contrast, memory-only setups tend to improve performance in \textsc{2048} and \textsc{Candy Crush}, especially for non-reasoning models that lack self-evaluation and thus repeat previous invalid moves. Overall, combining both modules yields generally higher gains, demonstrating the combined value of integrated perception and memory.

\begin{itemize}
  \item \textbf{General modular harness.}  
        We introduce a unified framework that decomposes an LLM/VLM agent into three plug‑and‑play components—perception, memory, and reasoning—enabling systematic investigation of each module’s role in multi‑turn decision‑making environments.
  \item \textbf{Game‑centric evaluation methodology.}  
        We leverage a diverse suite of Gym‑compatible titles (Sokoban, Candy Crush, 2048, Tetris) as a low‑barrier, scalable testbed; our adapter seamlessly integrates additional Gym or retro environments, harnessing existing game resources with minimal engineering effort.
  \item \textbf{Empirical validation and module analysis.}  
        Our experiments show consistent performance gains across diverse games and reveal distinct module contributions, demonstrating the general effectiveness of structured modular design.
\end{itemize}

\section{Related Works}

Games have long been foundational benchmarks in AI research, providing controlled environments to study planning and sequential decision-making — from early systems like TD-Gammon~\cite{tdgammon} to breakthroughs like AlphaGo~\cite{alphago}. This paradigm was further standardized by OpenAI Gym~\cite{towers2024gymnasium}, which introduced a unified interface for interacting with diverse game environments. Beyond classic chess and board game benchmarks, text-based games have also emerged as valuable benchmarks for evaluating agents’ language-driven reasoning and interaction capabilities. 
GameTraversalBenchmark\cite{nasir2024gametraversalbenchmark} focuses on diverse 2D grid-based game maps; GameBench\cite{costarelli2024gamebench} aims to assess strategy reasoning enhanced by advanced prompting techniques; GameEval\cite{qiao2023gameeval} centers on goal-driven conversational games where the dialogue process involves discussion, question answering, and voting; and GameArena\cite{hu2024gamearena} evaluates LLM reasoning through interactive gameplay with humans by having models guess or mention specific words. LMGame-Bench~\cite{hu2025lmgamebench} extends this trend by providing a quantitative interpretation of gaming performance across multiple tasks, analyzing action efficiency, success rates, and trajectory quality to assess reasoning and planning capabilities in LLM agents.
Separately, WhodunitBench~\cite{xie2024whodunitbench} targets multimodal agents, benchmarking perception, interaction, reasoning, and decision-making through murder mystery games.

Recent work has introduced various harnesses to improve the performance of gaming agents, such as perceptron, memory, and feedback mechanisms. 
Voyager~\cite{wang2023voyager} builds an embodied lifelong learning agent in Minecraft which refined skills from environmental feedback and reuse mastered skills through memory. 
The Factorio learning environment~\cite{hopkins2025factorio} introduces a harnessed agent setup that combines iterative program synthesis, symbolic memory, and environment feedback 
in a procedurally generated industrial simulation game.
Textgames~\cite{hudi2025textgames} maintains a history of responses and feedback, allowing models to reflect on their errors.
BALROG~\cite{paglieri2024balrog} evaluates games using textual inputs, with or without visual input, and provides observation history in prompts. However, it focuses on evaluating models without explicit thinking mechanisms, whereas our work includes recent reasoning models that show significant performance gains through ``thinking'' mechanisms. Similarly, LMAct~\cite{ruoss2024lmact} explores multimodal games with in-context imitation learning, emphasizing providing observation-action trajectories in prompts rather than providing environment feedback to prompt LLMs to reflect and improve from failure, as our harness does. 

Beyond game-focused benchmarks, a growing number of recent works have explored non-game agent settings that explicitly integrate reflection and perception mechanisms to enhance performance. Reflexion~\cite{shinn2023reflexion} introduces verbal reinforcement by prompting LLM agents to reflect on past errors and improve iteratively across tasks. ReAct~\cite{yao2023react} combines reasoning and acting through interleaved chain-of-thought generation and environment interaction, enabling more structured multi-step decision-making. AgentBench~\cite{liu2023agentbench} systematically evaluates LLM agents across multi-turn tasks, using history-augmented prompts and text-based action lists to test perception, reasoning, and decision-making in diverse domains. Systems like WebVoyager~\cite{he2024webvoyager} and AppAgent~\cite{zhang2025appagent} extend multi-turn agent research into real-world web and app environments, where agents reason over multimodal observations (e.g., screenshots, text) and select actions from structured lists (e.g., click, swipe, type) to complete tasks, demonstrating sophisticated perception and decision-making across sequential steps. Together, these works show the need for modular agent designs that help models reflect on their actions and adapt to what they see — principles we build into our game-focused system.
\section{Methods}
In designing our agent, we intentionally recycle well-known games not only for their familiarity and popularity but also because they encapsulate a broad spectrum of reasoning and interaction skills. Our goal is to preserve the original game settings that are carefully designed to challenge human cognition. In this section, we highlight the broad range of perception and generation abilities evaluated in our's game settings.

\begin{table*}[t]
\centering
\small
\caption{Model performance raw scores, evaluated both with and without harness. The reported results represent averages over three runs, except for models marked with *, which are based on a single run due to the high costs as of May 1, 2025.
}
\label{tab:model-game-performance-merged}
\vspace{0.5em}

\begin{tabular}{l|c|c|c|c|c}
\toprule
\textbf{Model} & \textbf{Harness} & Sokoban & Tetris & 2048 & Candy Crush \\
\midrule

{claude-3-5-sonnet-20241022} 
& No 
& 0.0{\scriptsize$\pm$0.0} 
& 12.3{\scriptsize$\pm$2.5} 
& 57.8{\scriptsize$\pm$16.4} 
& 17.0{\scriptsize$\pm$18.1}
\\
& Yes 
& 0.0{\scriptsize$\pm$0.0} 
& 14.7{\scriptsize$\pm$1.2} 
& 108.2{\scriptsize$\pm$5.8} 
& 106.0{\scriptsize$\pm$53.4} 
\\
\midrule
{claude-3-7-sonnet-20250219 (thinking)} 
& No  
& 0.0{\scriptsize$\pm$0.0} 
& 13.0{\scriptsize$\pm$0.0} 
& 114.2{\scriptsize$\pm$7.2} 
&126.3{\scriptsize$\pm$69.1} 
\\
& Yes 
& 2.3{\scriptsize$\pm$1.5} 
& 16.3{\scriptsize$\pm$2.3} 
& 113.3{\scriptsize$\pm$3.1} 
& 484.0{\scriptsize$\pm$53.7} 
\\
\midrule
{deepseek-r1} 
& No  
& N/A 
& N/A 
& N/A 
& N/A
\\
& Yes 
& 1.3{\scriptsize$\pm$1.2} 
& 14.3{\scriptsize$\pm$0.6} 
&105.2{\scriptsize$\pm$12.2}
&  447.3{\scriptsize$\pm$45.1} 
\\
\midrule
{gemini-2.5-flash-preview-04-17 (thinking)} 
& No  
& 0.0{\scriptsize$\pm$0.0} 
& 19.0{\scriptsize$\pm$4.6} 
& 107.4{\scriptsize$\pm$3.4} 
& 97.7{\scriptsize$\pm$36.1} 
\\
& Yes 
& 1.7{\scriptsize$\pm$1.5} 
& 16.3{\scriptsize$\pm$3.2} 
& 106.6{\scriptsize$\pm$5.3} 
&  334.7{\scriptsize$\pm$65.5} 
\\
\midrule
{gemini-2.5-pro-preview-05-06 (thinking)} 
& No  & 1.0{\scriptsize$\pm$0.0} &  12.3{\scriptsize$\pm$3.1} & 120.5{\scriptsize$\pm$3.9} & 177.3{\scriptsize$\pm$64.9} \\
                                         & Yes & 4.3{\scriptsize$\pm$0.6} & 23.3{\scriptsize$\pm$0.6} & 117.3{\scriptsize$\pm$5.9} & 416.3{\scriptsize$\pm$6.8} \\
\midrule
{grok-3-mini-beta (thinking)} 
  & No  
  & N/A 
  & N/A 
  & N/A 
  & N/A 
\\
  & Yes 
  & 5.7{\scriptsize$\pm$0.6} 
  & 21.3{\scriptsize$\pm$7.1} 
  & 118.6{\scriptsize$\pm$7.1} 
  & 254.0{\scriptsize$\pm$107.8} 
\\

\midrule
{llama-4-maverick-17b-128e-instruct-fp8} 
  & No  
    & 0.0{\scriptsize$\pm$0.0} 
    & 11.7{\scriptsize$\pm$1.2} 
    & 44.6{\scriptsize$\pm$11.8} 
    & 32.3{\scriptsize$\pm$41.4}  
\\
  & Yes 
    & 0.0{\scriptsize$\pm$0.0} 
    & 10.3{\scriptsize$\pm$1.5} 
    & 106.0{\scriptsize$\pm$3.8} 
    & 128.7{\scriptsize$\pm$57.2} 
\\

\midrule
{gpt-4.1-2025-04-14} 
  & No  
    & 0.0{\scriptsize$\pm$0.0} 
    & 13.0{\scriptsize$\pm$1.7} 
    & 94.5{\scriptsize$\pm$17.0} 
    & 101.0{\scriptsize$\pm$120.2} 
\\
  & Yes 
    & 0.0{\scriptsize$\pm$0.0} 
    & 13.7{\scriptsize$\pm$0.6} 
    & 105.7{\scriptsize$\pm$7.0} 
    & 182.0{\scriptsize$\pm$28.7} 
\\

\midrule
{gpt-4o-2024-11-20} 
  & No  
  & 0.0{\scriptsize$\pm$0.0} 
  & 14.7{\scriptsize$\pm$2.1} 
  & 70.4{\scriptsize$\pm$15.2} 
  & 59.0{\scriptsize$\pm$54.6} 
\\
  & Yes 
  & 0.0{\scriptsize$\pm$0.0} 
  & 14.0{\scriptsize$\pm$3.6} 
  & 106.7{\scriptsize$\pm$3.5} 
  & 147.3{\scriptsize$\pm$53.4} 
\\
\midrule
{o1-2024-12-17 *} 
  & No  
    & 0.0{\scriptsize$\pm$0.0} 
    & 13.0{\scriptsize$\pm$0.0} 
    & 128.1{\scriptsize$\pm$0.0} 
    & 90.0{\scriptsize$\pm$0.0}  
\\
  & Yes 
    & 2.3{\scriptsize$\pm$0.6} 
    & 35.0{\scriptsize$\pm$0.0} 
    & \textbf{128.9{\scriptsize$\pm$0.0}} 
    & 159.0{\scriptsize$\pm$0.0} 
\\

\midrule
{o1-mini-2024-09-12} 
  & No  
    & N/A 
    & N/A 
    & N/A 
    & N/A 
\\
  & Yes 
    & 1.3{\scriptsize$\pm$0.6} 
    & 11.7{\scriptsize$\pm$1.2} 
    & 114.0{\scriptsize$\pm$3.7} 
    & 48.0{\scriptsize$\pm$33.9}  
\\

\midrule
{o3-2025-04-16 *} 
  & No  
    & 2.0{\scriptsize$\pm$0.0} 
    & 31.0{\scriptsize$\pm$0.0} 
    & 128.2{\scriptsize$\pm$0.0} 
    & 106.0{\scriptsize$\pm$0.0} 
\\
  & Yes 
    & \textbf{8.0{\scriptsize$\pm$2.8} }
    & \textbf{42.0{\scriptsize$\pm$0.0}} 
    &128.0{\scriptsize$\pm$0.0} 
    & \textbf{647.0{\scriptsize$\pm$0.0} } 
\\

\midrule
{o4-mini-2025-04-16}
  & No  
    & 1.3{\scriptsize$\pm$0.6}
    & 15.0{\scriptsize$\pm$3.6}
    & 97.6{\scriptsize$\pm$29.2}
    & 110.7{\scriptsize$\pm$49.7}
\\
  & Yes 
    & 5.3{\scriptsize$\pm$1.2}
    & 25.3{\scriptsize$\pm$8.5}
    & 120.6{\scriptsize$\pm$4.9}
    & 487.3{\scriptsize$\pm$198.0}
\\

\midrule
Random & --
  & 0.0{\scriptsize$\pm$0.0}
  & 10.2{\scriptsize$\pm$1.8}
  & 100.4{\scriptsize$\pm$7.8}
  & 116.5{\scriptsize$\pm$51.5}
\\
\bottomrule
\end{tabular}%

\end{table*}

\subsection{Games}
\label{sec:game_design_rules}

\textbf{Tetris.} Tetris is a tile-matching puzzle game where players must strategically rotate and place falling Tetris tiles of 7 different geometric shapes to complete and clear horizontal lines. The game emphasizes (1) visual perception for pattern recognitions, (2) spatial reasoning for correct tile matching and geometric rotations~\cite{lau2017selective}, and (3) long-horizon planning with partial observability for decision-making on where and how to drop a tile~\cite{Demaine2003TetrisHard}. 

\textbf{Sokoban.} Sokoban is a grid-based puzzle game where the player pushes boxes to designated target locations within confined spaces. It emphasizes (1) visual perception, (2) spatial reasoning to navigate both the character and the box, and (3) long-horizon planning to avoid deadlocks~\cite{culberson1997sokoban}. The game’s low fault tolerance is especially pronounced. Many actions are irreversible, and a single wrong move can fail the puzzle.

\textbf{Candy Crush.} Candy Crush is a match-three puzzle game where players swap adjacent candies to form aligned sequences and trigger cascading effects to eliminate matched sequences. It requires (1) visual perception to identify different candies, (2) spatial reasoning to anticipate chain reactions at different locations, and (3) long-horizon planning to conserve moves to maximize total points. The gameplay features limited moves, making it crucial to plan moves carefully.

\textbf{2048.} 2048 is a sliding-tile puzzle game where players combine numbered tiles on a grid to reach the 2048 tile. It evaluates (1) visual perception for tracking tile values and movements, (2) spatial reasoning to manage merging paths, and (3) goal-directed planning to maximize merge potential~\cite{zaky2014minimax}. Errors compound quickly due to the game’s limited space and could lead to irreversible failure states.

\subsection{Game Metric Design}
\label{sec:game_metrics}

In line with Gymnasium, we treat a reward as a function
$  \mathcal{R}\colon \mathcal{S}\times\mathcal{A}\times\mathcal{S}\to\mathbb{R}$
which returns the payoff obtained when the agent executes action \(a\in\mathcal{A}\) in state \(s\in\mathcal{S}\) and transitions to state \(s'\). Our agent adopts this definition to evaluate agentic models and reports either progression rewards or long-horizon rewards.

\textbf{Progression rewards} provide dense, stepwise feedback as the agent makes forward advances. They apply to games with a primarily linear structure or no fixed endpoint, offering incremental scores that increase with each step—e.g., The running score in Tetris, 2048’s cumulative merge total, or Candy Crush’s cumulative eliminated candies.

\textbf{Long-horizon rewards} offer sparse credit awarded only upon completing a multi-step objective. These are common in games built around multi-step puzzles or multi-stage narratives, where rewards are granted after achieving the full goal—such as guiding a box through multiple steps to its target in \textsc{Sokoban}. Details in Appx.~\ref{appendix:game_details}.

\subsection{Modular Harness Design}
\label{sec:game_design_settings}
\textbf{Perception Modules.} Since video games are inherently multimodal, we design perception modules that convert UI inputs into textual representations of game states. We implement three modes: text, vision, and a combined mode. For grid-based games (\textsc{Sokoban}, \textsc{Candy Crush}, \textsc{2048}, \textsc{Tetris}), the text-based mode extracts the visual layout directly from the game backend and outputs a structured table listing object coordinates and properties—e.g., "Box at (2,3)", "Wall at (4,5)". This enables models to reason about spatial relationships without relying on raw image inputs, thereby minimizing perception errors. The vision-based mode leverages the perception capabilities of VLMs by converting rendered UI images into textual descriptions. To enhance accuracy, we overlay grid lines and coordinate labels on the image to reduce VLM perception errors. In the combined mode, we provide both the polished image and the deterministic textual representation from the backend, offering richer and more reliable input to the model.

\textbf{Memory Modules.}
Some games, like \textsc{2048}, require multi-step planning and the ability to correct invalid moves. Our memory module consists of two main components: (1) storing recent game trajectories and (2) reflecting on recent actions—serving as a form of short-term memory. Specifically, the module maintains the past $N$ game states and actions, and encourages the model to reflect on its last action by comparing the current state to previous ones. In line with Gymnasium, we treat this reflection process as an internal reward signal, where reward is defined as a function $\mathcal{R}\colon \mathcal{S} \times \mathcal{A} \times \mathcal{S} \to \mathbb{R}$. This self-evaluation helps the model recognize and correct suboptimal or invalid decisions. The goal of this module is to enhance coherent decision-making and adapt to the game's dynamic changes, enabling the agent to optimize its strategy rather than becoming stuck in repetitive or invalid moves.

\textbf{Reasoning Module.}  
The reasoning module serves as a general-purpose controller that integrates information from the perception and memory modules. It determines the agent’s final actions and allows flexible control over module integration—enabling us to activate or deactivate specific modules during evaluation. This design supports systematic analysis of how each component contributes to overall performance in multi-turn interactive environments.

\subsection{Prompt Standardization}
\label{sec:method_prompt_optimization}

\begin{algorithm}[tb]              
  \caption{DSPy Prompt Optimisation with SIMBA}
  \label{alg:dspy}
  \begin{algorithmic}[1]           
    \REQUIRE Training envs $\mathcal{E}_{\mathrm{train}}$; dev envs $\mathcal{E}_{\mathrm{dev}}$;
             target LMs $\mathcal{M}_t$; optimiser LMs
              $\mathcal{M}_o=\{\texttt{o3}\allowbreak,
                         \allowbreak\texttt{gemini-2.5}\allowbreak,
                         \allowbreak\texttt{claude-3.7}\allowbreak,
                         \allowbreak\texttt{deepseek-R1}\allowbreak,
                         \allowbreak\texttt{grok3-mini}\}$; max steps $k$
    \ENSURE  Best prompt template $\mathcal{P}^{\star}$

    \STATE $\mathcal{P}\leftarrow\textsc{ChainOfThought}(\text{``state}\!\rightarrow\!\text{action''})$
    \STATE $s_{\mathrm{best}}\leftarrow -\infty$
    \FORALL{$M_o\in\mathcal{M}_o$}
        \STATE \texttt{dspy.configure}$(M_o)$
        \STATE $\widehat{\mathcal{P}}\leftarrow
               \textsc{SIMBA}(\mathcal{M}_t,k)\!\cdot\!
               \texttt{compile}(\mathcal{P},\mathcal{E}_{\mathrm{train}})$
        \STATE $s_{\mathrm{avg}}\leftarrow 0$
        \FORALL{$M_t\in\mathcal{M}_t$}
            \STATE \texttt{dspy.configure}$(M_t)$
            \STATE $s_{\mathrm{avg}}\leftarrow s_{\mathrm{avg}}+
                   \texttt{Evaluate}(\widehat{\mathcal{P}},\mathcal{E}_{\mathrm{dev}})$
        \ENDFOR
        \STATE $s_{\mathrm{avg}}\leftarrow s_{\mathrm{avg}}/|\mathcal{M}_t|$
        \IF{$s_{\mathrm{avg}} > s_{\mathrm{best}}$}
            \STATE $s_{\mathrm{best}}\leftarrow s_{\mathrm{avg}}$;
                   $\mathcal{P}^{\star}\leftarrow\widehat{\mathcal{P}}$
        \ENDIF
    \ENDFOR
    \STATE \textsc{return} $\mathcal{P}^{\star}$
  \end{algorithmic}
\end{algorithm}


We design a two-stage prompt optimization framework to reduce performance variance and improve reliability across games. In the first stage, we apply empirical prompt engineering following recent agentic design practices~\cite{agashe2024agent, tan2024cradle}. Prompts are structured as [$\{\mathcal{J}_{[\min\left(0, i-N\right):i-1]}\}$, $R_{i-1}$, $s_i$], where $\mathcal{J}$ denotes a trajectory of prior states, actions, and rewards, and $R_{i-1}$ is a reflective summary generated by the memory module. 
In the second stage, we standardize optimization using DSPy~\cite{khattab2023dspy}. We employ SIMBA~\cite{wu2019simba} to iteratively refine each baseline prompt template across $k$ steps using cumulative reward as an objective. Optimization is bootstrapped using diverse large models, and the best-performing prompt $\mathcal{P}^{\star}$ is selected based on average performance across target LMs. Algorithm~\ref{alg:dspy} shows the DSPy optimization loop.

\section{Experiments}
In this section, we present the rankings of 13 state-of-the-art models, both with and without the modular harness, evaluated on a suite of 4 classical video games. We further evaluate how our modular harness improve models' performance across all games compared to unharness baseline and random gameplay in (Section ~\ref{sec:modular_harness_perf}). We also analyze the effectiveness of each harness module (Section ~\ref{sec:exp_rankings}). Finaly, through a case study of prompt standardization, we observe that prompts optimized through this bootstrapping process exhibit significantly lower performance variance compared to manually designed baselines, indicating increased robustness and standardization. (Section ~\ref{sec:system_design_prompt})

\begin{figure*}[h!]
  \centering
  \includegraphics[width=0.8\linewidth]{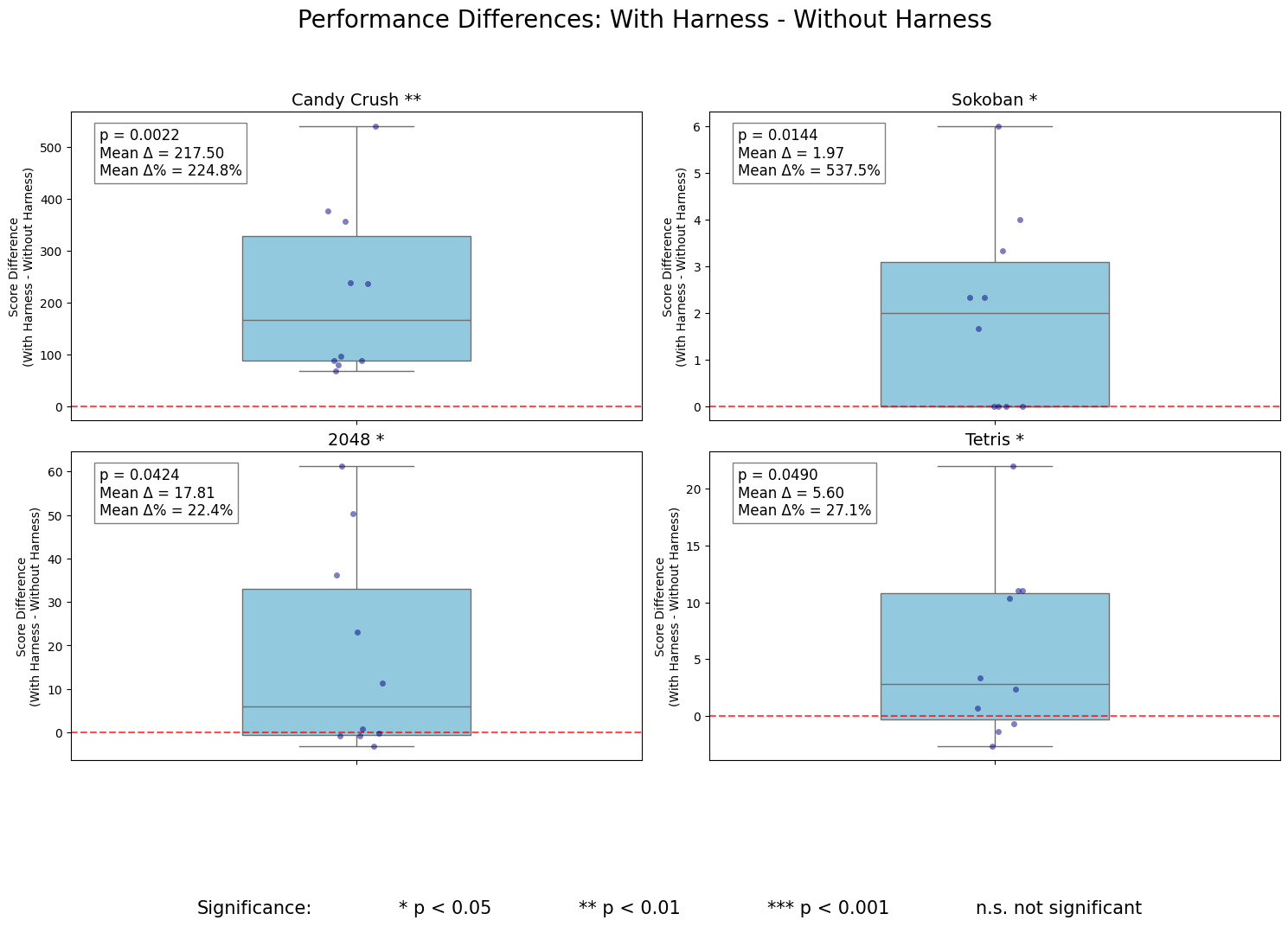}
  \caption{Distribution of paired score improvements (Harness – No Harness) across ten models for each game.}
  \label{fig:paired_diffs_boxplot}
  \vspace{-5pt}
\end{figure*}

\subsection{Modular Harness Performance: Quantitative Analysis}
\label{sec:modular_harness_perf}

To verify that our modular harness design leads to measurable improvements in model performance, we conduct a series of statistical analyses comparing harnessed, unharnessed, and random gameplay baselines. These results provide quantitative evidence that structured, modular control enhances decision-making in multi-turn interactive environments.

Due to the high cost of running latest foundation models, we report performance from a small-scale test setup: each model has three evaluation runs per game. While preliminary, these results already show a strong and consistent advantage for the modular harness design.

\subsubsection{Separation from Random Baseline: Glass’s $\delta$ Effect Sizes}
To quantify how far harnessed and unharnessed model outputs diverge from random play, we simulate 30 random gameplay runs per game to estimate the random baseline mean $\bar X_{\mathrm{rand}}$ and standard deviation $s_{\mathrm{rand}}$. We then compute Glass’s $\delta$ for each model–game–condition:

\begin{equation}
  \delta \;=\; \frac{\bar X_{\mathrm{model}} - \bar X_{\mathrm{rand}}}{s_{\mathrm{rand}}}
\label{eq:glass_delta}
\end{equation}

We exclude games like \textsc{Sokoban}, where random play exhibits zero variance, and focus on the remaining three games. 
Only one harnessed run yields a negative effect size, whereas 13 unharnessed runs do. Accordingly, 29 out of 30 harnessed $\delta$’s are positive, compared to 17 out of 30 for unharnessed. Moreover, the harnessed configuration outperforms the unharnessed one in 23 of 30 cases ($\approx76.7\%$). The average effect sizes remain:
\begin{equation*}
\begin{aligned}
  \bar\delta_{\mathrm{harness}} &= 2.757,  \\
  \bar\delta_{\mathrm{no}}      &= 0.009,  \\
  \Delta^*                      &= \bar\delta_{\mathrm{harness}} - \bar\delta_{\mathrm{no}}
                                  = 2.748.
\end{aligned}
\end{equation*}

These results show that our harnessed models achieve a much stronger departure from random behavior than the baseline counterparts.

\begin{table*}[h]
\centering
\small
\caption{Game scores of different models in \textit{Sokoban},  \textit{2048}, \textit{tetris}, and \textit{candy crush}under various conditions. ZS indicates zero-shot without any module support or memory prompt.}
\label{tab:combined_scores}
\begin{tabular}{lcccccc}
\toprule
\textbf{Model} & \textbf{Game} & \textbf{ZS} & \textbf{+Memory Only} & \textbf{+Perception Only} & \textbf{+Both} \\
\midrule
o4‑mini-2025‑04‑16           & Sokoban & 1.3{\scriptsize$\pm$0.6} & 1.3{\scriptsize$\pm$0.6} & 5.3{\scriptsize$\pm$2.1} & 5.3{\scriptsize$\pm$1.2} \\
gemini‑2.5-Pro-03‑25         & & 1.0{\scriptsize$\pm$0.0} & 1.0{\scriptsize$\pm$0.0} & 6.0{\scriptsize$\pm$2.0} & 4.3{\scriptsize$\pm$0.6} \\
claude‑3-7-Sonnet            &  & 0.0{\scriptsize$\pm$0.0} & 0.3{\scriptsize$\pm$0.6} & 0.7{\scriptsize$\pm$0.6} & 2.3{\scriptsize$\pm$1.5} \\
llama-4-maverick             & & 0.0{\scriptsize$\pm$0.0} & 0.0{\scriptsize$\pm$0.0} & 0.0{\scriptsize$\pm$0.0} & 0.0{\scriptsize$\pm$0.0} \\
claude-3-5-sonnet            & & 0.0{\scriptsize$\pm$0.0} & 0.0{\scriptsize$\pm$0.0} & 0.0{\scriptsize$\pm$0.0} & 0.0{\scriptsize$\pm$0.0} \\
gpt-4o-2024-11-20            & & 0.0{\scriptsize$\pm$0.0} & 0.0{\scriptsize$\pm$0.0} & 0.0{\scriptsize$\pm$0.0} & 0.0{\scriptsize$\pm$0.0} \\
\midrule
o4‑mini-2025‑04‑16           & 2048 & 97.6{\scriptsize$\pm$29.2} & 115.1{\scriptsize$\pm$9.7} & 117.0{\scriptsize$\pm$6.4} & 120.6{\scriptsize$\pm$4.9} \\
gemini‑2.5-Pro-03‑25         &  & 120.5{\scriptsize$\pm$3.9} & 118.0{\scriptsize$\pm$8.5} & 117.4{\scriptsize$\pm$5.8} & 117.3{\scriptsize$\pm$5.9} \\
claude‑3-7-Sonnet            &  & 114.2{\scriptsize$\pm$7.2} & 107.1{\scriptsize$\pm$5.1} & 115.3{\scriptsize$\pm$2.3} & 113.3{\scriptsize$\pm$3.1} \\
llama-4-maverick             & & 44.6{\scriptsize$\pm$11.8} & 98.1{\scriptsize$\pm$3.8} & 73.7{\scriptsize$\pm$15.6} & 106.0{\scriptsize$\pm$3.8} \\
claude-3-5-sonnet            & & 57.8{\scriptsize$\pm$16.4} & 102.5{\scriptsize$\pm$1.6} & 66.3{\scriptsize$\pm$9.6} & 108.2{\scriptsize$\pm$5.8} \\
gpt-4o-2024-11-20            & & 70.4{\scriptsize$\pm$15.2} & 107.0{\scriptsize$\pm$6.3} & 73.3{\scriptsize$\pm$5.4} & 106.7{\scriptsize$\pm$3.5} \\
\midrule
o4‑mini-2025‑04‑16           & Tetris & 15.0{\scriptsize$\pm$3.6} 
& 14.7{\scriptsize$\pm$3.2} & 38.0{\scriptsize$\pm$11.3} & 25.3{\scriptsize$\pm$8.5} \\
gemini‑2.5-Pro-03‑25         &  & 12.3{\scriptsize$\pm$3.1} & 
15.0{\scriptsize$\pm$1.0} & 21.3{\scriptsize$\pm$3.5} & 23.3{\scriptsize$\pm$0.6} \\
claude‑3-7-Sonnet            &  & 13.0{\scriptsize$\pm$0.0}  & 15.7{\scriptsize$\pm$3.1} &  20.0{\scriptsize$\pm$3.6} & 16.3{\scriptsize$\pm$2.3} \\
llama-4-maverick             &  & 11.7{\scriptsize$\pm$1.2}  & 10.3{\scriptsize$\pm$1.5} &  8.7{\scriptsize$\pm$1.5}& 10.3{\scriptsize$\pm$1.5} \\
claude-3-5-sonnet            &  & 12.3{\scriptsize$\pm$2.5} & 13.3{\scriptsize$\pm$4.7} &  14.3{\scriptsize$\pm$3.1} & 14.7{\scriptsize$\pm$1.2} \\
gpt-4o-2024-11-20            &  & 14.7{\scriptsize$\pm$2.1} & 14.0{\scriptsize$\pm$2.0}  &  18.0{\scriptsize$\pm$6.6}& 106.7{\scriptsize$\pm$3.5} \\
\midrule
o4‑mini-2025‑04‑16           & Candy Crush &  110.7{\scriptsize$\pm$49.7} & 202.3{\scriptsize$\pm$88.0} & 32{\scriptsize$\pm$3.5} & 487.3{\scriptsize$\pm$198.0} \\
gemini‑2.5-Pro-03‑25         &  & 177.3{\scriptsize$\pm$64.9} & 93.7{\scriptsize$\pm$58.4} & 386.7{\scriptsize$\pm$138.5} & 416.3{\scriptsize$\pm$6.8} \\
claude‑3-7-Sonnet            &  & 126.3{\scriptsize$\pm$69.1} & 187.3{\scriptsize$\pm$151.6} & 270.3{\scriptsize$\pm$240.0} & 484.0{\scriptsize$\pm$53.7} \\
llama-4-maverick             &  & 32.3{\scriptsize$\pm$41.4} & 123.3{\scriptsize$\pm$83.9} & 110.0{\scriptsize$\pm$23.4} &  128.7{\scriptsize$\pm$57.2} \\
claude-3-5-sonnet            &  &  17.0{\scriptsize$\pm$18.1} & 120.3{\scriptsize$\pm$41.5} & 10.7{\scriptsize$\pm$8.6} & 106.0{\scriptsize$\pm$53.4}  \\
gpt-4o-2024-11-20            &  & 59.0{\scriptsize$\pm$54.6} & 49.3{\scriptsize$\pm$38.4} & 78.3{\scriptsize$\pm$24.7} & 147.3{\scriptsize$\pm$53.4} \\
\bottomrule
\end{tabular}
\end{table*}

\subsubsection{Direct Comparison: Paired‑Sample t‑Tests}
To further confirm performance improvements, we perform paired-sample $t$-tests, comparing harnessed and unharnessed scores for each model across all four games to test that whether modular harness design significantly improves unharness design.

\textsc{Candy Crush} (+217.50 points, \(t(9)=4.22,\,p=0.0022\)), \textsc{Sokoban} (+1.97, \(t(9)=3.02,\,p=0.0144\)), \textsc{2048} (+17.81, \(t(9)=2.36,\,p=0.0424\)), and \textsc{Tetris} (+5.60, \(t(9)=2.27,\,p=0.0490\))—the improvements are statistically significant (\(p < 0.05\)). 

Figure~\ref{fig:paired_diffs_boxplot} shows the per-model score differences (Harness – No Harness) for each game. \textsc{Candy Crush} show the highest median gains.

\subsection{Modular Ablation}
\label{sec:exp_rankings}
We design different levels of scaffolding, as described in Section \ref{sec:game_design_settings}, to better differentiate model capabilities. Results in Table~\ref{tab:combined_scores} show that the harness leads to consistent and sometimes substantial gains across both games.

\textbf{Perception Modules.}  
In \textit{Sokoban}, perception modules such as vision scaffolding lead to meaningful performance gains, particularly for models like Gemini and O4-mini. These structured spatial inputs appear to unlock planning behaviors that are otherwise latent under raw token-based input. A similar pattern emerges in \textit{Tetris}, another spatially grounded task: perception provides a strong boost to models across the board, with O4-mini showing the most notable improvement. These findings underscore the importance of perceptual grounding for environments with geometric or spatial dynamics.

\textbf{Memory Modules.}  
Memory proves particularly impactful in temporally extended games. In \textit{2048}, models with weak zero-shot performance, such as Claude-3.5 and GPT-4o, improve substantially with memory support, highlighting its importance in long-horizon planning. The effect is even more dramatic in \textit{Candy Crush}, a game with complex temporal dependencies and delayed rewards. Memory not only raises average scores but also reduces variance, stabilizing performance across episodes. This reinforces memory's role in preserving context and strategy over time.

\textbf{Combined Support.}  
The strongest gains occur when both modules are enabled. Across all four games, combined perception and memory support produces additive or even multiplicative improvements. For example, in \textit{Candy Crush}, Gemini and Claude-3.7 achieve massive gains under the full support condition, revealing performance distinctions that are otherwise obscured. Notably, even top-performing models diverge in this regime, suggesting that the dual-module setting serves as a higher-resolution benchmark. It exposes nuanced strengths and weaknesses that zero-shot evaluations or single-module setups may fail to reveal.

\textbf{Summary.}  
Perception and memory provide complementary advantages across different game domains. Perception is most beneficial in spatially structured environments like \textit{Sokoban} and \textit{Tetris}, whereas memory is crucial for games requiring long-term planning, such as \textit{2048} and \textit{Candy Crush}. Together, these modules enhance not only raw performance but also the discriminative power of evaluations---making them essential tools for rigorous benchmarking.

\begin{table*}[!t]
  \centering
  \small
  \caption{Glass’s \(\delta\) per Model, Condition, and Game (rounded to 3 decimals).  
  Positive \(\delta\) means performance is pulled away from the random baseline.}
  \label{tab:glass_delta_full}
  \begin{tabular}{llrrr}
    \toprule
    \textbf{Model} & \textbf{Cond.} & \textbf{2048} & \textbf{Candy Crush} & \textbf{Tetris} \\
    \midrule
    {claude-3-5-sonnet-20241022}  
      & With    &  0.992 & –0.204 &  2.524 \\
      & Without & –5.446 & –1.933 &  1.215 \\
    \addlinespace
    {claude-3-7-sonnet-20250219 (thinking)}  
      & With    &  1.648 &  7.140 &  3.459 \\
      & Without &  1.752 &  0.191 &  1.589 \\
    \addlinespace
    {gemini-2.5-flash-preview-04-17 (thinking)}  
      & With    &  0.787 &  4.238 &  3.459 \\
      & Without &  0.883 & –0.366 &  4.955 \\
    \addlinespace
    {gemini-2.5-pro-preview-05-06 (thinking)}  
      & With    &  2.148 &  5.825 &  7.386 \\
      & Without &  2.558 &  1.182 &  1.215 \\
    \addlinespace
    {gpt-4.1-2025-04-14}  
      & With    &  0.675 &  1.273 &  1.963 \\
      & Without & –0.762 & –0.301 &  1.589 \\
    \addlinespace
    {gpt-4o-2024-11-20}  
      & With    &  0.793 &  0.599 &  2.150 \\
      & Without & –3.833 & –1.117 &  2.524 \\
    \addlinespace
    {llama-4-maverick-17b-128e-instruct-fp8}  
      & With    &  0.707 &  0.236 &  0.093 \\
      & Without & –7.124 & –1.635 &  0.841 \\
    \addlinespace
    {o1-2024-12-17}  
      & With    &  3.631 &  0.826 & 13.930 \\
      & Without &  3.530 & –0.515 &  1.589 \\
    \addlinespace
    {o3-2025-04-16}  
      & With    &  3.516 & 10.306 & 17.856 \\
      & Without &  3.541 & –0.204 & 11.686 \\
    \addlinespace
    {o4-mini-2025-04-16}  
      & With    &  2.577 &  7.204 &  8.508 \\
      & Without & –0.368 & –0.113 &  2.711 \\
    \bottomrule
  \end{tabular}
\end{table*}
 
\subsection{Prompt Standardization}
\label{sec:system_design_prompt}
Prompt engineering is known to significantly affect LLM performance in interactive environments~\cite{wang25lvlmgame}. However, we find that even well-crafted empirical prompts can result in performance variability exceeding one standard deviation (Table~\ref{tab:prompt_optimization}). To address this, we evaluate our two-stage prompt optimization method (Section~\ref{sec:method_prompt_optimization}), which combines empirical design with DSPy-based refinement. Results show that DSPy optimization consistently improves average performance and reduces variance across models and games. Detailed examples are provided in Appendix~\ref{appendix:prompt_optimization}.

\subsection{Correlation Study}
\label{sec:exp_analysis}
This section investigates whether performance on our harness-enabled game suite serves as a proxy for broader model capabilities. Specifically, we examine whether game performance correlates with scores on widely used LLM benchmarks targeting reasoning, vision, and planning.

We evaluated the following 8 models: Claude-3.5-Sonnet, Claude-3.7-Sonnet-Thinking, Gemini-2.5-Pro-Preview, Llama-4-Maverick, GPT-4o, o1, o3, and o4-mini in 20 established benchmarks that span factual knowledge~\cite{vals2025mmlu, MMLU_paper, phan2024humanity, hle_leaderboard, hle_textonly_leaderboard, rein2023gpqa, gpqa_leaderboard}, physics~\cite{qiu2025phybench, EMMA_benchmark}, mathematics~\cite{EMMA_benchmark, math500_vals_leaderboard, patel2024aime, aime25_vals_leaderboard, livebench, livebench_leaderboard}, code generation~\cite{zhuo2024bigcodebench, aider_leaderboards, bigcodebench_leaderboard}, visual reasoning~\cite{vista_leaderboard, chatbot_arena_leaderboard, yue2023mmmu, mmmu_vals_leaderboard}, language understanding~\cite{sirdeshmukh2025multichallenge, multichallenge_leaderboard}, and puzzle solving~\cite{wang2025enigmaeval, enigmaeval_leaderboard, mazur2025nytconnections} to analyze correlations with game-based performance. See Appx.~\ref{appendix:correlation} for the complete benchmark list. 

Fig.~\ref{fig:corr_only} reveals positive correlations among commonly evaluated benchmarks and games. Sokoban shows strong correlations with math and coding benchmarks. Tetris and 2048 closely align with pattern recognition tasks like EnigmaEval and NYT-connections. Candy Crush is notably related to coding, hinting at algorithmic reasoning.

\section{Conclusion}
We introduce a general modular harness design that consistently improves model performance in multi-turn interactive environments. Our methodology leverages the abundance of existing games as testbeds, enabling systematic investigation of general-purpose agent design. By ablating individual modules—perception, memory, and reasoning—we can isolate their contributions and evaluate their impact on overall agent behavior. Furthermore, our framework supports a wide range of diverse game environments, providing a scalable and reproducible solution for studying modular agents at scale.

\bibliography{paper}

\begin{thebibliography}{70}
\providecommand{\natexlab}[1]{#1}
\providecommand{\url}[1]{\texttt{#1}}
\expandafter\ifx\csname urlstyle\endcsname\relax
  \providecommand{\doi}[1]{doi: #1}\else
  \providecommand{\doi}{doi: \begingroup \urlstyle{rm}\Url}\fi

\bibitem[Abbasi \& Saffidine(2018)Abbasi and Saffidine]{2048Hard2018}
Abbasi, A. and Saffidine, A.
\newblock Threes!, 1024!, and 2048 are hard.
\newblock \emph{Theoretical Computer Science}, 2018.

\bibitem[Abdulhai et~al.(2023)Abdulhai, White, Snell, Sun, Hong, Zhai, Xu, and Levine]{abdulhai2023lmrl}
Abdulhai, M., White, I., Snell, C., Sun, C., Hong, J., Zhai, Y., Xu, K., and Levine, S.
\newblock Lmrl gym: Benchmarks for multi-turn reinforcement learning with language models.
\newblock \emph{arXiv preprint arXiv:2311.18232}, 2023.

\bibitem[Agashe et~al.(2024)Agashe, Han, Gan, Yang, Li, and Wang]{agashe2024agent}
Agashe, S., Han, J., Gan, S., Yang, J., Li, A., and Wang, X.~E.
\newblock Agent s: An open agentic framework that uses computers like a human, 2024.
\newblock URL \url{https://arxiv.org/abs/2410.08164}.

\bibitem[{Aider Team}()]{aider_leaderboards}
{Aider Team}.
\newblock Aider llm leaderboards.
\newblock \url{https://aider.chat/docs/leaderboards/}.
\newblock Accessed: 2025-05-14.

\bibitem[Besta et~al.(2024)Besta, Blach, Kubicek, et~al.]{besta2024got}
Besta, M., Blach, N., Kubicek, A., et~al.
\newblock Graph of thoughts: Solving elaborate problems with large language models.
\newblock \emph{arXiv preprint arXiv:2308.09687}, 2024.

\bibitem[{BigCodeBench Team}()]{bigcodebench_leaderboard}
{BigCodeBench Team}.
\newblock Bigcodebench leaderboard.
\newblock \url{https://bigcode-bench.github.io/}.
\newblock Accessed: 2025-05-14.

\bibitem[Costarelli et~al.(2024)Costarelli, Allen, Hauksson, Sodunke, Hariharan, Cheng, Li, Clymer, and Yadav]{costarelli2024gamebench}
Costarelli, A., Allen, M., Hauksson, R., Sodunke, G., Hariharan, S., Cheng, C., Li, W., Clymer, J., and Yadav, A.
\newblock Gamebench: Evaluating strategic reasoning abilities of llm agents.
\newblock \emph{arXiv preprint arXiv:2406.06613}, 2024.

\bibitem[Culberson(1997)]{culberson1997sokoban}
Culberson, J.
\newblock Sokoban is pspace-complete.
\newblock 1997.

\bibitem[Demaine et~al.(2003)Demaine, Hohenberger, and Liben‑Nowell]{Demaine2003TetrisHard}
Demaine, E.~D., Hohenberger, S., and Liben‑Nowell, D.
\newblock Tetris is hard, even to approximate.
\newblock In \emph{International Computing and Combinatorics Conference (COCOON)}, volume 2697 of \emph{Lecture Notes in Computer Science}, pp.\  351--363. Springer, 2003.
\newblock URL \url{https://doi.org/10.1007/3-540-45127-7_34}.

\bibitem[Deng et~al.(2023)Deng, Gu, Zheng, et~al.]{deng2023mind2web}
Deng, X., Gu, Y., Zheng, B., et~al.
\newblock Mind2web: Towards a generalist agent for the web.
\newblock \emph{arXiv preprint arXiv:2308.00957}, 2023.

\bibitem[Deng et~al.(2024)Deng, Zheng, Chen, et~al.]{webarena2024}
Deng, Y., Zheng, B., Chen, S., et~al.
\newblock Webarena: A realistic web environment for building autonomous agents.
\newblock \emph{arXiv preprint arXiv:2307.13854}, 2024.

\bibitem[Dor \& Zwick(1999)Dor and Zwick]{SokobanNPhard1999}
Dor, D. and Zwick, U.
\newblock Motion planning with obstacles: {Sokoban} is np-hard.
\newblock \emph{Computational Geometry}, 1999.

\bibitem[Hao et~al.(2025)Hao, Gu, Wang, Li, Yang, Wang, and Cheng]{EMMA_benchmark}
Hao, Y., Gu, J., Wang, H.~W., Li, L., Yang, Z., Wang, L., and Cheng, Y.
\newblock Can mllms reason in multimodality? emma: An enhanced multimodal reasoning benchmark.
\newblock \emph{arXiv preprint arXiv:2501.05444}, 2025.
\newblock URL \url{https://arxiv.org/abs/2501.05444}.

\bibitem[He et~al.(2024)He, Yao, Ma, Yu, Dai, Zhang, Lan, and Yu]{he2024webvoyager}
He, H., Yao, W., Ma, K., Yu, W., Dai, Y., Zhang, H., Lan, Z., and Yu, D.
\newblock Webvoyager: Building an end-to-end web agent with large multimodal models.
\newblock \emph{arXiv preprint arXiv:2401.13919}, 2024.

\bibitem[Hendrycks et~al.(2021)Hendrycks, Burns, Kadavath, Arora, Basart, Tang, et~al.]{MMLU_paper}
Hendrycks, D., Burns, C., Kadavath, S., Arora, P., Basart, S., Tang, D.~S., et~al.
\newblock Measuring massive multitask language understanding.
\newblock \emph{arXiv preprint arXiv:2009.03300}, 2021.
\newblock URL \url{https://arxiv.org/abs/2009.03300}.

\bibitem[Hopkins et~al.(2025)Hopkins, Bakler, and Khan]{hopkins2025factorio}
Hopkins, J., Bakler, M., and Khan, A.
\newblock Factorio learning environment.
\newblock \emph{arXiv preprint arXiv:2503.09617}, 2025.

\bibitem[Hu et~al.(2024)Hu, Li, Xie, Jiang, Stoica, Jin, and Zhang]{hu2024gamearena}
Hu, L., Li, Q., Xie, A., Jiang, N., Stoica, I., Jin, H., and Zhang, H.
\newblock Gamearena: Evaluating llm reasoning through live computer games.
\newblock \emph{arXiv preprint arXiv:2412.06394}, 2024.

\bibitem[Hu et~al.(2025)Hu, Huo, Zhang, Yu, Xing, Stoica, Rosing, Jin, and Zhang]{hu2025lmgamebench}
Hu, L., Huo, M., Zhang, Y., Yu, H., Xing, E.~P., Stoica, I., Rosing, T., Jin, H., and Zhang, H.
\newblock Lmgame-bench: How good are llms at playing games?
\newblock \emph{arXiv preprint arXiv:2505.15146}, 2025.

\bibitem[Hudi et~al.(2025)Hudi, Winata, Zhang, and Aji]{hudi2025textgames}
Hudi, F., Winata, G.~I., Zhang, R., and Aji, A.~F.
\newblock Textgames: Learning to self-play text-based puzzle games via language model reasoning.
\newblock \emph{arXiv preprint arXiv:2502.18431}, 2025.

\bibitem[Khattab et~al.(2023)Khattab, Singhvi, Maheshwari, Zhang, Santhanam, Vardhamanan, Haq, Sharma, Joshi, Moazam, Miller, Zaharia, and Potts]{khattab2023dspy}
Khattab, O., Singhvi, A., Maheshwari, P., Zhang, Z., Santhanam, K., Vardhamanan, S., Haq, S., Sharma, A., Joshi, T.~T., Moazam, H., Miller, H., Zaharia, M., and Potts, C.
\newblock Dspy: Compiling declarative language model calls into self-improving pipelines.
\newblock \emph{arXiv preprint arXiv:2310.03714}, 2023.

\bibitem[Lau-Zhu et~al.(2017)Lau-Zhu, Holmes, Butterfield, and Holmes]{lau2017selective}
Lau-Zhu, A., Holmes, E.~A., Butterfield, S., and Holmes, J.
\newblock Selective association between tetris game play and visuospatial working memory: A preliminary investigation.
\newblock \emph{Applied cognitive psychology}, 31\penalty0 (4):\penalty0 438--445, 2017.

\bibitem[Liu et~al.(2023)Liu, Yu, Zhang, Xu, Lei, Lai, Gu, Ding, Men, Yang, et~al.]{liu2023agentbench}
Liu, X., Yu, H., Zhang, H., Xu, Y., Lei, X., Lai, H., Gu, Y., Ding, H., Men, K., Yang, K., et~al.
\newblock Agentbench: Evaluating llms as agents.
\newblock \emph{arXiv preprint arXiv:2308.03688}, 2023.

\bibitem[{LiveBench Team}()]{livebench_leaderboard}
{LiveBench Team}.
\newblock Livebench leaderboard.
\newblock \url{https://livebench.ai/#/?Coding=a&Mathematics=a&Data+Analysis=a&Language=a&IF=a}.
\newblock Accessed: 2025-05-14.

\bibitem[{LMSYS Org}()]{chatbot_arena_leaderboard}
{LMSYS Org}.
\newblock Chatbot arena leaderboard.
\newblock \url{https://lmarena.ai/leaderboard}.
\newblock Accessed: 2025-05-14.

\bibitem[Mazur()]{mazur2025nytconnections}
Mazur, L.
\newblock Nyt connections benchmark: Evaluating llms with extended word association puzzles.
\newblock \url{https://github.com/lechmazur/nyt-connections}.
\newblock Accessed: 2025-05-14.

\bibitem[Nasir et~al.(2024)Nasir, James, and Togelius]{nasir2024gametraversalbenchmark}
Nasir, M.~U., James, S., and Togelius, J.
\newblock Gametraversalbenchmark: Evaluating planning abilities of large language models through traversing 2d game maps.
\newblock In \emph{The Thirty-eight Conference on Neural Information Processing Systems Datasets and Benchmarks Track}, 2024.
\newblock URL \url{https://openreview.net/forum?id=sAxVIWQOzo}.

\bibitem[Newell(1990)]{NewellUTC1990}
Newell, A.
\newblock \emph{Unified Theories of Cognition}.
\newblock Harvard University Press, 1990.

\bibitem[Paglieri et~al.(2024)Paglieri, Cupia{\l}, Coward, Piterbarg, Wolczyk, Khan, Pignatelli, Kuci{\'n}ski, Pinto, Fergus, et~al.]{paglieri2024balrog}
Paglieri, D., Cupia{\l}, B., Coward, S., Piterbarg, U., Wolczyk, M., Khan, A., Pignatelli, E., Kuci{\'n}ski, {\L}., Pinto, L., Fergus, R., et~al.
\newblock Balrog: Benchmarking agentic llm and vlm reasoning on games.
\newblock \emph{arXiv preprint arXiv:2411.13543}, 2024.

\bibitem[Patel et~al.(2024)Patel, Chakraborty, Suttle, Wang, Bedi, and Manocha]{patel2024aime}
Patel, B., Chakraborty, S., Suttle, W.~A., Wang, M., Bedi, A.~S., and Manocha, D.
\newblock Aime: Ai system optimization via multiple llm evaluators.
\newblock \emph{arXiv preprint arXiv:2410.03131}, 2024.
\newblock URL \url{https://arxiv.org/abs/2410.03131}.

\bibitem[Phan et~al.(2024)Phan, Gatti, Han, Li, Hu, Zhang, Zhang, Shaaban, Ling, Shi, Choi, Agrawal, Chopra, Khoja, Kim, Ren, Hausenloy, Zhang, Mazeika, Yue, Wang, and Hendrycks]{phan2024humanity}
Phan, L., Gatti, A., Han, Z., Li, N., Hu, J., Zhang, H., Zhang, C. B.~C., Shaaban, M., Ling, J., Shi, S., Choi, M., Agrawal, A., Chopra, A., Khoja, A., Kim, R., Ren, R., Hausenloy, J., Zhang, O., Mazeika, M., Yue, S., Wang, A., and Hendrycks, D.
\newblock Humanity’s last exam: Benchmarking ai on grade school science olympiad exams.
\newblock \emph{arXiv preprint arXiv:2501.14249}, 2024.
\newblock URL \url{https://arxiv.org/abs/2501.14249}.

\bibitem[Qiao et~al.(2023)Qiao, Wu, Liang, Li, and Duan]{qiao2023gameeval}
Qiao, D., Wu, C., Liang, Y., Li, J., and Duan, N.
\newblock Gameeval: Evaluating llms on conversational games.
\newblock \emph{arXiv preprint arXiv:2308.10032}, 2023.

\bibitem[Qiu et~al.(2025)Qiu, Guo, Song, Sun, Cai, Wei, Luo, Yin, Zhang, Hu, Wang, Tang, Chang, Liu, Zhou, Zhang, Zhang, Liu, Li, Zhang, Jing, Yin, Ren, Fu, Ji, Wang, Tian, Lv, Man, Li, Tao, Sun, Liang, Mu, Li, Zhang, Zhang, Li, Xia, Lin, Shen, Chen, Xiong, Wang, Wang, Ni, Zhang, Cui, Shao, Cao, Luo, Yang, Zhang, and Zhu]{qiu2025phybench}
Qiu, S., Guo, S., Song, Z.-Y., Sun, Y., Cai, Z., Wei, J., Luo, T., Yin, Y., Zhang, H., Hu, Y., Wang, C., Tang, C., Chang, H., Liu, Q., Zhou, Z., Zhang, T., Zhang, J., Liu, Z., Li, M., Zhang, Y., Jing, B., Yin, X., Ren, Y., Fu, Z., Ji, J., Wang, W., Tian, X., Lv, A., Man, L., Li, J., Tao, F., Sun, Q., Liang, Z., Mu, Y., Li, Z., Zhang, J.-J., Zhang, S., Li, X., Xia, X., Lin, J., Shen, Z., Chen, J., Xiong, Q., Wang, B., Wang, F., Ni, Z., Zhang, B., Cui, F., Shao, C., Cao, Q.-H., Luo, M.-X., Yang, Y., Zhang, M., and Zhu, H.~X.
\newblock Phybench: Holistic evaluation of physical perception and reasoning in large language models.
\newblock \emph{arXiv preprint arXiv:2504.16074}, 2025.
\newblock URL \url{https://arxiv.org/abs/2504.16074}.

\bibitem[Rein et~al.(2023)Rein, Hou, Stickland, Petty, Pang, Dirani, Michael, and Bowman]{rein2023gpqa}
Rein, D., Hou, B.~L., Stickland, A.~C., Petty, J., Pang, R.~Y., Dirani, J., Michael, J., and Bowman, S.~R.
\newblock Gpqa: Graded physics question answering benchmark for large language models.
\newblock \emph{arXiv preprint arXiv:2311.12022}, 2023.
\newblock URL \url{https://arxiv.org/abs/2311.12022}.

\bibitem[Research(2024)]{osworld2024}
Research, O.
\newblock Osworld: Benchmarking multimodal agents for open-ended computer tasks.
\newblock \url{https://os-world.github.io/}, 2024.

\bibitem[Ruoss et~al.(2024)Ruoss, Pardo, Chan, Li, Mnih, and Genewein]{ruoss2024lmact}
Ruoss, A., Pardo, F., Chan, H., Li, B., Mnih, V., and Genewein, T.
\newblock Lmact: A benchmark for in-context imitation learning with long multimodal demonstrations.
\newblock \emph{arXiv preprint arXiv:2412.01441}, 2024.

\bibitem[{Scale AI}({\natexlab{a}})]{enigmaeval_leaderboard}
{Scale AI}.
\newblock Enigmaeval benchmark leaderboard.
\newblock \url{https://scale.com/leaderboard/enigma_eval}, {\natexlab{a}}.
\newblock Accessed: 2025-05-14.

\bibitem[{Scale AI}({\natexlab{b}})]{hle_leaderboard}
{Scale AI}.
\newblock Humanity's last exam leaderboard.
\newblock \url{https://scale.com/leaderboard/humanitys_last_exam}, {\natexlab{b}}.
\newblock Accessed: 2025-05-14.

\bibitem[{Scale AI}({\natexlab{c}})]{hle_textonly_leaderboard}
{Scale AI}.
\newblock Humanity's last exam leaderboard (text only).
\newblock \url{https://scale.com/leaderboard/humanitys_last_exam_text_only}, {\natexlab{c}}.
\newblock Accessed: 2025-05-14.

\bibitem[{Scale AI}({\natexlab{d}})]{multichallenge_leaderboard}
{Scale AI}.
\newblock Multichallenge leaderboard.
\newblock \url{https://scale.com/leaderboard/multichallenge}, {\natexlab{d}}.
\newblock Accessed: 2025-05-14.

\bibitem[{Scale AI}({\natexlab{e}})]{vista_leaderboard}
{Scale AI}.
\newblock Vista: Visual language understanding benchmark leaderboard.
\newblock \url{https://scale.com/leaderboard/visual_language_understanding}, {\natexlab{e}}.
\newblock Accessed: 2025-05-14.

\bibitem[Shen et~al.(2024)Shen, Jain, Xiao, Amlekar, Hadji, Podolny, and Talwalkar]{shen2024scribeagent}
Shen, J., Jain, A., Xiao, Z., Amlekar, I., Hadji, M., Podolny, A., and Talwalkar, A.
\newblock Scribeagent: Towards specialized web agents using production-scale workflow data.
\newblock \emph{arXiv preprint arXiv:2411.15004}, 2024.
\newblock Published November 22, 2024.

\bibitem[Shinn et~al.(2023)Shinn, Cassano, Gopinath, Narasimhan, and Yao]{shinn2023reflexion}
Shinn, N., Cassano, F., Gopinath, A., Narasimhan, K., and Yao, S.
\newblock Reflexion: Language agents with verbal reinforcement learning.
\newblock \emph{Advances in Neural Information Processing Systems}, 36:\penalty0 8634--8652, 2023.

\bibitem[Silver et~al.(2016)Silver, Huang, Maddison, Guez, Sifre, Van Den~Driessche, Schrittwieser, Antonoglou, Panneershelvam, Lanctot, et~al.]{alphago}
Silver, D., Huang, A., Maddison, C.~J., Guez, A., Sifre, L., Van Den~Driessche, G., Schrittwieser, J., Antonoglou, I., Panneershelvam, V., Lanctot, M., et~al.
\newblock Mastering the game of go with deep neural networks and tree search.
\newblock \emph{nature}, 529\penalty0 (7587):\penalty0 484--489, 2016.

\bibitem[Sirdeshmukh et~al.(2025)Sirdeshmukh, Deshpande, Mols, Jin, Cardona, Lee, Kritz, Primack, Yue, and Xing]{sirdeshmukh2025multichallenge}
Sirdeshmukh, V., Deshpande, K., Mols, J., Jin, L., Cardona, E.-Y., Lee, D., Kritz, J., Primack, W., Yue, S., and Xing, C.
\newblock Multichallenge: A realistic multi-turn conversation evaluation benchmark challenging to frontier llms.
\newblock \emph{arXiv preprint arXiv:2501.17399}, 2025.
\newblock URL \url{https://arxiv.org/abs/2501.17399}.

\bibitem[Tan et~al.(2024)Tan, Zhang, Xu, et~al.]{tan2024cradle}
Tan, W., Zhang, W., Xu, X., et~al.
\newblock Cradle: Empowering foundation agents towards general computer control.
\newblock \emph{arXiv preprint arXiv:2403.03186}, 2024.

\bibitem[Tesauro(1994)]{tdgammon}
Tesauro, G.
\newblock Td-gammon, a self-teaching backgammon program, achieves master-level play.
\newblock \emph{Neural computation}, 6\penalty0 (2):\penalty0 215--219, 1994.

\bibitem[Towers et~al.(2024)Towers, Kwiatkowski, Terry, Balis, De~Cola, Deleu, Goul{\~a}o, Kallinteris, Krimmel, Wierman, et~al.]{towers2024gymnasium}
Towers, M., Kwiatkowski, A., Terry, J., Balis, J.~U., De~Cola, G., Deleu, T., Goul{\~a}o, M., Kallinteris, A., Krimmel, M., Wierman, A., et~al.
\newblock Gymnasium: A standard interface for reinforcement learning environments.
\newblock \emph{arXiv preprint arXiv:2407.17032}, 2024.
\newblock URL \url{https://arxiv.org/abs/2407.17032}.

\bibitem[{Vals AI}()]{aime25_vals_leaderboard}
{Vals AI}.
\newblock Aime benchmark leaderboard.
\newblock \url{https://www.vals.ai/benchmarks/aime-2025-05-09}.
\newblock Accessed: 2025-05-14.

\bibitem[{VALS AI}()]{gpqa_leaderboard}
{VALS AI}.
\newblock Gpqa benchmark leaderboard.
\newblock \url{https://www.vals.ai/benchmarks/gpqa-05-09-2025}.
\newblock Accessed: 2025-05-14.

\bibitem[{Vals AI}({\natexlab{a}})]{math500_vals_leaderboard}
{Vals AI}.
\newblock Math 500 benchmark leaderboard.
\newblock \url{https://www.vals.ai/benchmarks/math500-05-09-2025}, {\natexlab{a}}.
\newblock Accessed: 2025-05-14.

\bibitem[{Vals AI}({\natexlab{b}})]{mmmu_vals_leaderboard}
{Vals AI}.
\newblock Mmmu benchmark leaderboard.
\newblock \url{https://www.vals.ai/benchmarks/mmmu-05-09-2025}, {\natexlab{b}}.
\newblock Accessed: 2025-05-14.

\bibitem[{VALS AI}(2025)]{vals2025mmlu}
{VALS AI}.
\newblock Mmlu-pro benchmark leaderboard.
\newblock \url{https://www.vals.ai/benchmarks/mmlu_pro-05-09-2025}, 2025.
\newblock Accessed: 2025-05-13.

\bibitem[Vlasov \& Chase(2024)Vlasov and Chase]{langgraph2024}
Vlasov, R. and Chase, H.
\newblock Langgraph: Stateful orchestration framework for agentic workflows.
\newblock \url{https://www.langchain.com/langgraph}, 2024.

\bibitem[Walsh(2014)]{CandyCrushNPhard2014}
Walsh, T.
\newblock Candy crush is np-hard.
\newblock \emph{arXiv preprint arXiv:1403.1911}, 2014.

\bibitem[Wang et~al.(2025{\natexlab{a}})Wang, Lee, Menghini, Mols, Doughty, Khoja, Lynch, Hendryx, Yue, and Hendrycks]{wang2025enigmaeval}
Wang, C.~J., Lee, D., Menghini, C., Mols, J., Doughty, J., Khoja, A., Lynch, J., Hendryx, S., Yue, S., and Hendrycks, D.
\newblock Enigmaeval: A benchmark of long multimodal reasoning challenges.
\newblock \emph{arXiv preprint arXiv:2502.08859}, 2025{\natexlab{a}}.
\newblock URL \url{https://arxiv.org/abs/2502.08859}.

\bibitem[Wang et~al.(2023)Wang, Xie, Jiang, Mandlekar, Xiao, Zhu, Fan, and Anandkumar]{wang2023voyager}
Wang, G., Xie, Y., Jiang, Y., Mandlekar, A., Xiao, C., Zhu, Y., Fan, L., and Anandkumar, A.
\newblock Voyager: An open-ended embodied agent with large language models.
\newblock \emph{arXiv preprint arXiv:2305.16291}, 2023.

\bibitem[Wang et~al.()Wang, Zhuang, and Wu]{wang25lvlmgame}
Wang, X., Zhuang, B., and Wu, Q.
\newblock Are large vision language models good game players?
\newblock In \emph{The Thirteenth International Conference on Learning Representations}.

\bibitem[Wang et~al.(2024)Wang, Mao, Fried, and Neubig]{wang2024awm}
Wang, Z., Mao, J., Fried, D., and Neubig, G.
\newblock Agent workflow memory.
\newblock \emph{arXiv preprint arXiv:2409.07429}, 2024.

\bibitem[Wang et~al.(2025{\natexlab{b}})Wang, Wang, Wang, Zhang, Li, Yang, Yu, Nguyen, Liu, Gottlieb, et~al.]{wang2025ragen}
Wang, Z., Wang, K., Wang, Q., Zhang, P., Li, L., Yang, Z., Yu, K., Nguyen, M.~N., Liu, L., Gottlieb, E., et~al.
\newblock Ragen: Understanding self-evolution in llm agents via multi-turn reinforcement learning.
\newblock \emph{arXiv preprint arXiv:2504.20073}, 2025{\natexlab{b}}.

\bibitem[White et~al.(2024)White, Dooley, Roberts, Pal, Feuer, Jain, Shwartz-Ziv, Jain, Saifullah, Naidu, Hegde, LeCun, Goldstein, Neiswanger, and Goldblum]{livebench}
White, C., Dooley, S., Roberts, M., Pal, A., Feuer, B., Jain, S., Shwartz-Ziv, R., Jain, N., Saifullah, K., Naidu, S., Hegde, C., LeCun, Y., Goldstein, T., Neiswanger, W., and Goldblum, M.
\newblock Livebench: A challenging, contamination-free llm benchmark.
\newblock \emph{arXiv preprint arXiv:2406.19314}, 2024.
\newblock URL \url{https://arxiv.org/abs/2406.19314}.

\bibitem[Wu et~al.(2019)Wu, Sun, Matlock, Liu, Tian, and Kamilov]{wu2019simba}
Wu, Z., Sun, Y., Matlock, A., Liu, J., Tian, L., and Kamilov, U.~S.
\newblock {SIMBA}: Scalable inversion in optical tomography using deep denoising priors.
\newblock \emph{arXiv preprint arXiv:1911.13241}, 2019.
\newblock URL \url{https://arxiv.org/abs/1911.13241}.

\bibitem[Xie et~al.(2024)Xie, Zhang, Chen, Wan, and Li]{xie2024whodunitbench}
Xie, J., Zhang, R., Chen, Z., Wan, X., and Li, G.
\newblock Whodunitbench: Evaluating large multimodal agents via murder mystery games.
\newblock In \emph{The Thirty-eight Conference on Neural Information Processing Systems Datasets and Benchmarks Track}, 2024.
\newblock URL \url{https://openreview.net/forum?id=qmvtDIfbmS}.

\bibitem[Yang et~al.(2023)Yang, Yue, and He]{yang2023autogpt}
Yang, H., Yue, S., and He, Y.
\newblock Autogpt for online decision making: Benchmarks and additional opinions.
\newblock \emph{arXiv preprint arXiv:2306.02224}, 2023.

\bibitem[Yao et~al.(2023)Yao, Zhao, Yu, Du, Shafran, Narasimhan, and Cao]{yao2023react}
Yao, S., Zhao, J., Yu, D., Du, N., Shafran, I., Narasimhan, K., and Cao, Y.
\newblock React: Synergizing reasoning and acting in language models.
\newblock In \emph{International Conference on Learning Representations (ICLR)}, 2023.

\bibitem[Yue et~al.(2023)Yue, Ni, Zhang, Zheng, Liu, Zhang, Stevens, Jiang, Ren, Sun, et~al.]{yue2023mmmu}
Yue, X., Ni, Y., Zhang, K., Zheng, T., Liu, R., Zhang, G., Stevens, S., Jiang, D., Ren, W., Sun, Y., et~al.
\newblock Mmmu: A massive multi-discipline multimodal understanding and reasoning benchmark for expert agi.
\newblock \emph{arXiv preprint arXiv:2311.16502}, 2023.
\newblock URL \url{https://arxiv.org/abs/2311.16502}.

\bibitem[Zaky(2014)]{zaky2014minimax}
Zaky, A.
\newblock Minimax and expectimax algorithm to solve 2048.
\newblock 2014.

\bibitem[Zhai et~al.(2024)Zhai, Bai, Lin, Pan, Tong, Zhou, Suhr, Xie, LeCun, Ma, and Levine]{NEURIPS2024rl4vlm}
Zhai, Y., Bai, H., Lin, Z., Pan, J., Tong, S., Zhou, Y., Suhr, A., Xie, S., LeCun, Y., Ma, Y., and Levine, S.
\newblock Fine-tuning large vision-language models as decision-making agents via reinforcement learning.
\newblock In Globerson, A., Mackey, L., Belgrave, D., Fan, A., Paquet, U., Tomczak, J., and Zhang, C. (eds.), \emph{Advances in Neural Information Processing Systems}, volume~37, pp.\  110935--110971. Curran Associates, Inc., 2024.
\newblock URL \url{https://proceedings.neurips.cc/paper_files/paper/2024/file/c848b7d3adc08fcd0bf1df3101ba6728-Paper-Conference.pdf}.

\bibitem[Zhang et~al.(2025)Zhang, Yang, Liu, Li, Han, Chen, Huang, Fu, and Yu]{zhang2025appagent}
Zhang, C., Yang, Z., Liu, J., Li, Y., Han, Y., Chen, X., Huang, Z., Fu, B., and Yu, G.
\newblock Appagent: Multimodal agents as smartphone users.
\newblock In \emph{Proceedings of the 2025 CHI Conference on Human Factors in Computing Systems}, pp.\  1--20, 2025.

\bibitem[Zhou et~al.(2025)Zhou, Jiang, Tian, Weston, Levine, Sukhbaatar, and Li]{zhou2025sweet}
Zhou, Y., Jiang, S., Tian, Y., Weston, J., Levine, S., Sukhbaatar, S., and Li, X.
\newblock Sweet-rl: Training multi-turn llm agents on collaborative reasoning tasks.
\newblock \emph{arXiv preprint arXiv:2503.15478}, 2025.

\bibitem[Zhuo et~al.(2024)Zhuo, Vu, Chim, Hu, Yu, Widyasari, Yusuf, Zhan, He, Paul, Brunner, Gong, Hoang, Zebaze, Hong, Li, Kaddour, Xu, Zhang, Yadav, Jain, Gu, Cheng, Liu, Liu, Wang, Lo, Hui, Muennighoff, Fried, Du, de~Vries, and Von~Werra]{zhuo2024bigcodebench}
Zhuo, T.~Y., Vu, M.~C., Chim, J., Hu, H., Yu, W., Widyasari, R., Yusuf, I. N.~B., Zhan, H., He, J., Paul, I., Brunner, S., Gong, C., Hoang, T., Zebaze, A.~R., Hong, X., Li, W.-D., Kaddour, J., Xu, M., Zhang, Z., Yadav, P., Jain, N., Gu, A., Cheng, Z., Liu, J., Liu, Q., Wang, Z., Lo, D., Hui, B., Muennighoff, N., Fried, D., Du, X., de~Vries, H., and Von~Werra, L.
\newblock Bigcodebench: Benchmarking code generation with diverse function calls and complex instructions.
\newblock \emph{arXiv preprint arXiv:2406.15877}, 2024.
\newblock URL \url{https://arxiv.org/abs/2406.15877}.

\end{thebibliography}
\bibliographystyle{icml2025}

\newpage
\appendix
\onecolumn
\section{Game Setting Details}
\label{appendix:game_details}
Because games are well-designed, we utilize their built-in metrics to quantify models' proficiency. For each game, we choose the single score that most faithfully reflects a model’s capability, then transform and normalize it onto a continuous, linear scale. This curation ensures that our evaluation can sensitively capture performance differences and supports consistent statistical analysis.
\begin{itemize}
  \item \textbf{Sokoban:} Total number of boxes pushed onto targets, summed over all levels, until the first deadlock.
  \item \textbf{Tetris:} Total pieces dropped plus total lines cleared, measured up to the point of game over.
  \item \textbf{2048:} Sum of all merged tile values (e.g.\ merging two 2’s yields +2), recorded until the board stagnates (no merges or moves that change the board for ten consecutive turns). We then report
    \[
      \text{Score}_{2048} \;=\; 10 \times \log_{2}\!\bigl(\text{total merged sum}\bigr).
    \]
  \item \textbf{Candy Crush:} Total number of candies eliminated over a fixed 50–move session.
\end{itemize}

\section{Harness Analsysi}
\subsection{Quantitive Study}

\subsection{Qualitative Study}
\label{appendix:qualitative_analysis}

In this section, we present case studies that demonstrate the reasoning traces of the latest models during their 2048 gameplay.

\subsubsection{Memory Module}

\textbf{Memory Module Reflection Comparison: GPT-4o vs. o3} 

In our 2048 case study, GPT-4o's reflections tend to focus on immediate effect of each move ("up increased empty spaces but produced no merges"), whereas o3's reflections unpack exactly which tiles merged, how merge ladders are forming, and even quantify changes in empty-cell counts("two 2's became a 4 and two 16's became 32's, opening an extra cell"). Importantly, an interactive game environment isn't a one-off Q\&A but a multi-turn dialogue with a constantly shifting game board - each action affects next decision. o3 better links its fine-grained observations to concrete next steps, integrating short-term move outcomes into its long-term corner-dominance strategy. The success of local decisions is crucial to achieving long‑term goals in a multi‑turn interactive game environment.

\noindent\textbf{Highlight Legend:}

\hlred{Red text}: the model’s \emph{immediate} game‐state reflection  
    
\hlyellow{Yellow text}: o3’s \emph{immediate} game‐state reflection  
    
\hlgreen{Green text}: the model’s \emph{future move} suggestion  

\tcbset{
  reasoningstyle/.style={
     colback=gray!5,
     colframe=purple,          
     coltitle=green,             
     fonttitle=\bfseries,      
     boxrule=1.2pt,   
     arc=4pt,              
     left=1em,right=1em,top=1em,bottom=1em,
  }
}

\newtcolorbox{reasoningbox-4o}{
  reasoningstyle,
  title={GPT-4o Game 2048 Memory Module Reflections}
}

\newtcolorbox{reasoningbox-o3}{
  reasoningstyle,
  title={o3 Game 2048 Memory Module Reflections}
}
\tcbset{%
  sampleboxstyle/.style={%
    reasoningstyle,
    boxrule=0.8pt,
    coltitle=white,
    fonttitle=\bfseries,
    leftrule=0pt, rightrule=0pt, bottomrule=0pt, toprule=0pt,
    arc=2pt,
    left=1em, right=1em, top=1em, bottom=1em,
    boxsep=2pt,
    colback=white
  }
}

\newtcolorbox{samplebox}[2][]{%
  sampleboxstyle,
  title={Sample~#2},
  #1
}
\begin{reasoningbox-4o}
  \begin{samplebox}{1}
    \begin{minipage}[c]{0.48\textwidth}
      \textbf{Reflection:} \hlred{The move "up" increased empty spaces from 2 to 4, improving board flexibility while maintaining the highest tile (256) in position. However, no significant merges occurred, suggesting the move was minimally effective for progress.} The key pattern is the consistent maintenance of the highest tile in a corner, aligning with the strategy. \hlgreen{Future moves should focus on creating merge opportunities, particularly around the 256 tile, while avoiding disruptions to its placement. Consider prioritizing "down" or "left" next to consolidate lower tiles toward the main corner, optimizing for larger merges.}
    \end{minipage}\hfill
    \begin{minipage}[c]{0.48\textwidth}
      \centering
      \includegraphics[width=0.9\linewidth]{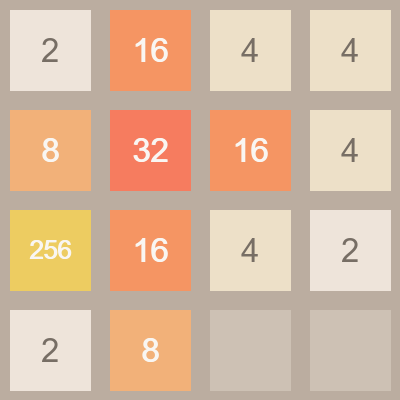}
    \end{minipage}
  \end{samplebox}

  \begin{samplebox}{2}
    \begin{minipage}[c]{0.48\textwidth}
      \textbf{Reflection:} \hlred{The "left" action was suboptimal as it failed to create merges or improve board position, leaving only two empty spaces while slightly disrupting alignment.} The highest tile (256) remains safe in a corner, but the addition of a 2 at (0,3) weakens the merging path potential. Merges were not prioritized, and the action did not capitalize on available opportunities to consolidate tiles. \hlgreen{Future moves should focus on maximizing merges and preserving space, particularly by aligning higher tiles on one edge to maintain control and avoid corner disruption. Aim to optimize empty space usage for better tile positioning.}
    \end{minipage}\hfill
    \begin{minipage}[c]{0.48\textwidth}
      \centering
      \includegraphics[width=0.9\linewidth]{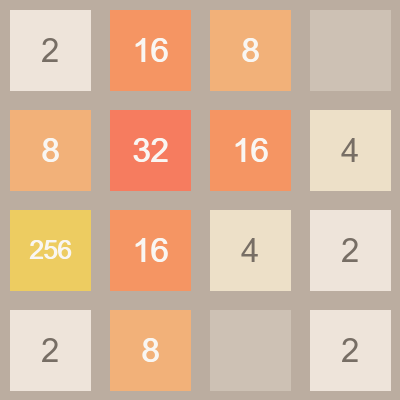}
    \end{minipage}
  \end{samplebox}

  \begin{samplebox}{3}
    \begin{minipage}[c]{0.48\textwidth}
      \textbf{Reflection:} \hlred{The "down" move consolidated smaller tiles, creating merges (e.g., 2s to 4s), but did not impact the highest tile (256) or significantly improve positioning.} Empty spaces shifted, but the same two remain, limiting opportunities. The action maintained balance but lacked strategic progression toward higher merges. The key issue is the scattered mid-tier tiles, which could restrict future merges. \hlgreen{Focus on better clustering tiles toward one corner and maintain merge paths for the 256 tile. Prioritizing "up" or "left" next may help consolidate the board and preserve the highest tile’s strategic position.}
    \end{minipage}\hfill
    \begin{minipage}[c]{0.48\textwidth}
      \centering
      \includegraphics[width=0.9\linewidth]{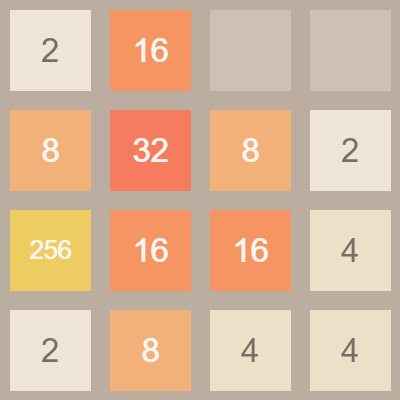}
    \end{minipage}
  \end{samplebox}
\end{reasoningbox-4o}

\begin{reasoningbox-o3}
  \begin{samplebox}{1}
    \begin{minipage}[c]{0.48\textwidth}
      \textbf{Reflection:} \hlyellow{The right swipe merged the two 8‑tiles in row 2 (3 → 4), created a 16 and opened an extra gap; the new 2 spawned harmlessly at (1,0).} 256 stayed anchored in the bottom‑right, and the right column is now 256‑128‑64‑16, a strong merge ladder. Empty count stayed at four, so the board is still breathable.Good move: it scored, preserved the corner strategy and improved board order.Watch for: isolated 2’s creeping into the left half; they can freeze rows if ignored. \hlgreen{Next: keep sliding Right/Down to compress toward the bottom‑right; avoid Up unless desperate to prevent dislodging the 256.}

    \end{minipage}\hfill
    \begin{minipage}[c]{0.48\textwidth}
      \centering
      \includegraphics[width=0.9\linewidth]{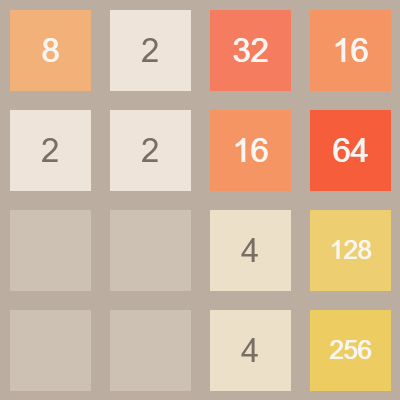}
    \end{minipage}
  \end{samplebox}

  \vspace{1em}

  \begin{samplebox}{2}
    \begin{minipage}[c]{0.48\textwidth}
      \textbf{Reflection:} \hlyellow{Up combined the two 2‑tiles in column 2 into an 8 and merged the pair of 2‑tiles in column 1 into a 4, netting +2 empty squares; after the new spawn you still gained one extra space (4 → 5).} 256 stayed anchored in the bottom‑right, so the corner strategy is intact. Good move: it advanced tile values, improved board looseness and preserved ordering. Watch the freshly spawned 2 at (3,2); if the bottom row clogs you’ll lose the 256 corner. \hlgreen{Prefer more up/left sequences to clear column 2 and funnel new merges toward the right column.}
    \end{minipage}\hfill
    \begin{minipage}[c]{0.48\textwidth}
      \centering
      \includegraphics[width=0.9\linewidth]{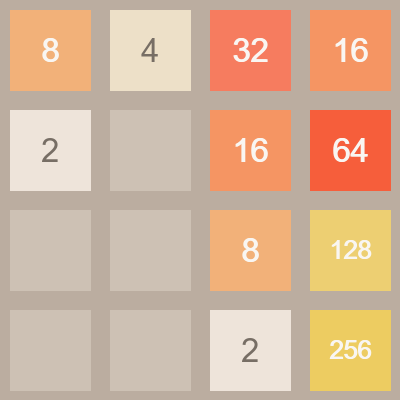}
    \end{minipage}
  \end{samplebox}

  \vspace{1em}

  \begin{samplebox}{3}
    \begin{minipage}[c]{0.48\textwidth}
      \textbf{Reflection:} \hlyellow{Down kept the 256 anchored and slid every column smoothly, but it yielded zero merges and cost one empty slot (5 → 4).} Column 0 merely shuffled 8‑2 downward and spawned a new 2 on top, so no pairing potential was created. Repeated space‑losing moves like this can clog the board; always look for turns that both respect the corner strategy and create or set up merges (e.g., left could have joined the two 2‑tiles on the bottom row next). \hlgreen{Prioritize moves that grow space or prepare obvious combines before defaulting to safe ‘corner‑holding’ slides.}
    \end{minipage}\hfill
    \begin{minipage}[c]{0.48\textwidth}
      \centering
      \includegraphics[width=0.9\linewidth]{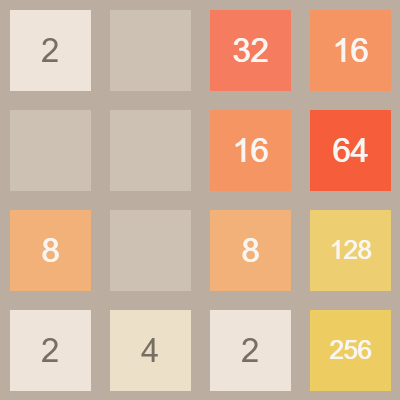}
    \end{minipage}
  \end{samplebox}
\end{reasoningbox-o3}

\textbf{Memory Module Prompt Template} 
\newtcolorbox{memory-prompt}{
  reasoningstyle,
  title={Game 2048 Memory Module}
}
\begin{memory-prompt}
\textbf{\textcolor{blue}{system\_prompt:}}

You are an analytical assistant for a 2048 AI agent. \\
Your task is to generate a brief, insightful reflection on the game state changes and the effectiveness of recent actions.Focus on strategic insights and patterns that would help the agent make better decisions.\\
Keep your reflections short, precise, and actionable.\\

\textbf{\textcolor{blue}{user\_prompt:}}

Please analyze the following 2048 game states and actions to generate a brief reflection:\\
Previous Game States and Actions:\\
\textcolor{blue}{\{Previous Game History\}}\\
Focus your reflection on:\\
1. How the game state changed after the last action\\
2. Whether the action was effective for the situation\\
3. Patterns or issues to be aware of\\
4. Any strategic insights for future actions\\
Keep your reflection under 100 words and focus only on the most important insights.
\end{memory-prompt}

\section{Benchmark List for Correlation Study}
\label{appendix:correlation}

We use 20 publicly available benchmarks spanning seven capability categories, including factual knowledge, physics, mathematics, coding, visual reasoning, language understanding, and puzzle solving. These benchmarks are chosen to provide a comprehensive view of general-purpose model abilities and to support the correlation and decomposition analyses.

Table~\ref{tab:benchmark_ranks_only} summarizes the per-model rankings across all benchmarks, grouped by category. The rankings are used to compute Spearman correlations and to uncover latent capability axes through low-rank decomposition.

\begin{figure}[t]
    \centering
    \includegraphics[width=0.5\linewidth]{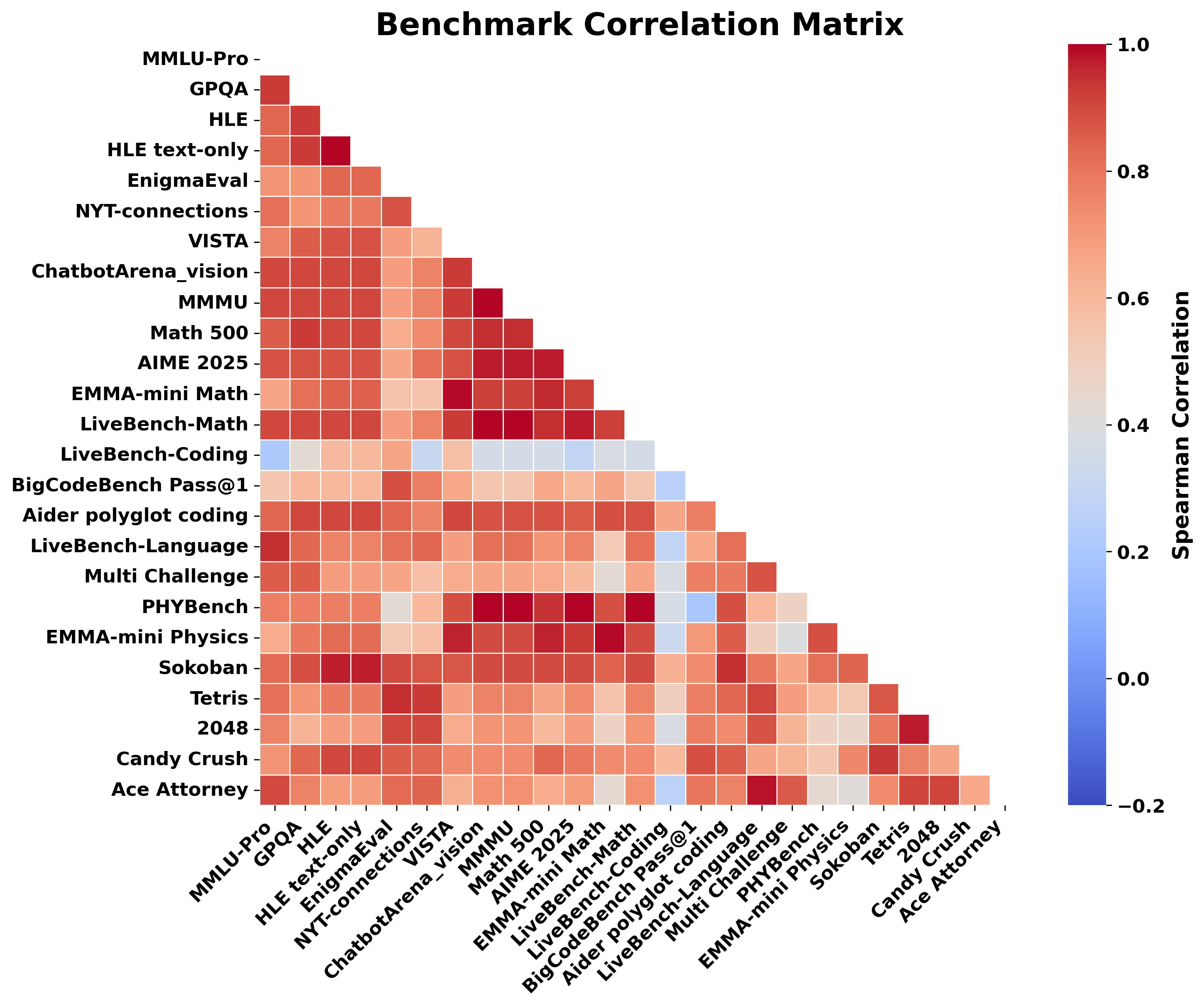}
    \caption{Spearman correlation among our benchmark and other benchmarks.}
    \label{fig:corr_only}
\end{figure}

\label{appendix:benchmark_ranks_only}

\begin{table}[h]
\centering
\small
\caption{
Model rankings (1 = best) across 20 benchmarks, grouped by capability category. Abbreviated model names: C3.5 = claude-3.5-Sonnet-20241022, C3.7 = claude-3.7-Sonnet-20250219-thinking, G4O = gpt-4o-2024-11-20, O1 = o1-2024-12-17, O3 = o3-2025-04-16, Gem = gemini-2.5-pro-preview-05-06(thinking), L4 = llama-4-maverick-17b-128e-instruct-fp8, O4m = o4-mini.}
\begin{tabular}{llcccccccc}
\toprule
\textbf{Category} & \textbf{Benchmark} & C3.5 & C3.7 & G4O & O1 & O3 & Gem & L4 & O4m \\
\midrule
{Factual}
  & MMLU-Pro       & 7 & 4 & 2 & 6 & 8 & 3 & 1 & 5 \\
  & GPQA           & 7 & 3 & 2 & 6 & 8 & 5 & 1 & 4 \\
  & HLE            & 7 & 4 & 3 & 6 & 8 & 5 & 1 & 2 \\
  & HLE (Text)     & 7 & 4 & 3 & 6 & 8 & 5 & 1 & 2 \\
\midrule
{Physics}
  & EMMA-Physics   & 7 & 4 & 1 & 8 & 6 & 5 & 3 & 2 \\
  & PHYBench       & 7 & 5 & 1 & 8 & 6 & 4 & 2 & 3 \\
\midrule
{Math}
  & Math 500       & 8 & 4 & 1 & 6 & 7 & 5 & 2 & 3 \\
  & AIME 2025      & 8 & 5 & 1 & 6 & 7 & 4 & 2 & 3 \\
  & EMMA-Math      & 6 & 4 & 1 & 8 & 6 & 5 & 3 & 2 \\
  & LiveBench-Math & 7 & 5 & 1 & 6 & 8 & 4 & 2 & 3 \\
\midrule
{Code}
  & BigCodeBench   & 5 & 1 & 3 & 6 & 4 & 2 & 7 & 8 \\
  & Aider Coding   & 6 & 4 & 2 & 8 & 7 & 5 & 1 & 3 \\
  & LiveBench-Code & 3 & 4 & 5 & 8 & 6 & 7 & 2 & 1 \\
\midrule
{Vision}
  & VISTA          & 6 & 4 & 1 & 7 & 8 & 5 & 3 & 2 \\
  & MMMU           & 7 & 5 & 1 & 6 & 8 & 4 & 2 & 3 \\
  & Chatbot Arena (Vision) & 7 & 5 & 1 & 6 & 8 & 4 & 2 & 3 \\
\midrule
{Language}
  & MultiChallenge & 5 & 2 & 3 & 7 & 8 & 4 & 1 & 6 \\
  & LiveBench-Lang & 6 & 4 & 3 & 7 & 8 & 2 & 1 & 5 \\
\midrule
{Puzzle}
  & EnigmaEval     & 6 & 4 & 5 & 8 & 7 & 3 & 1 & 2 \\
  & NYT Connections& 8 & 5 & 4 & 7 & 6 & 2 & 1 & 3 \\
\bottomrule
\end{tabular}
\vspace{0.5em}
\label{tab:benchmark_ranks_only}
\end{table}

\section{Prompt Optimization}
\label{appendix:prompt_optimization}
\subsection{Empirically Optimized Baseline Prompts}


\tcbset{
  reasoningstyle/.style={
     colback=gray!5,
     colframe=purple,          
     coltitle=green,             
     fonttitle=\bfseries,      
     boxrule=1.2pt,   
     arc=4pt,              
     left=1em,right=1em,top=1em,bottom=1em,
  }
}

\newtcolorbox{vision-prompt-1}{
  reasoningstyle,
  title={Game 2048 — Empirically Optimized Prompt Template 1}
}

\newtcolorbox{vision-prompt-2}{
  reasoningstyle,
  title={Game 2048 — Empirically Optimized Prompt Template 2}
}

\newtcolorbox{vision-prompt-dspy1}{
  reasoningstyle,
  title={Game 2048 — DSPy Optimized Prompt Template 1}
}

\newtcolorbox{vision-prompt-dspy2}{
  reasoningstyle,
  title={Game 2048 — DSPy Optimized Prompt Template 2}
}

\begin{vision-prompt-1}

\textbf{\textcolor{blue}{system\_prompt:}}\\
You are an intelligent AI player playing the 2048 game. Your goal is to make strategic moves to combine tiles and reach the highest possible tile value.\\

IMPORTANT: You MUST format your response using EXACTLY these lines:\\
thought: [Your reasoning about the game state]\\
move: [move]\\
Where [move] must be one of: "up", "down", "left", or "right".\\
Do not include \# or any other prefix. Start directly with "thought:" followed by your analysis.\\

\textbf{\textcolor{blue}{user\_prompt:}}\\
2048 Game Quick Guide:\\
Primary Goal: Combine like tiles to create tiles with higher values.\\
Ultimate Goal: Create a tile with the value 2048 or higher.\\
Game Mechanics:\\
- The game is played on a 4x4 grid.\\
- Each move (up, down, left, right) shifts all tiles in that direction.\\
- Tiles with the same value that collide during a move combine into a single tile with twice the value.\\
- After each move, a new tile (2 or 4) appears in a random empty cell.\\
- The game ends when there are no valid moves left.\\
Action Space:\\
You must select one of these 4 moves:\\
- up: Shift all tiles upward\\
- down: Shift all tiles downward\\
- left: Shift all tiles to the left\\
- right: Shift all tiles to the right\\
Key Strategies:\\
1. Build a stable structure - Keep your highest value tiles in a corner.\\
2. Maintain a clear path - Always have a direction where you can combine tiles.\\
3. Chain reactions - Set up sequences of merges that can happen in a single move.\\
4. Look ahead - Think about the consequences of your moves 2-3 steps ahead.\\
5. Building patterns - Common patterns include: (1) Snake/Zig-zag pattern: Arrange tiles in decreasing order in a zigzag; (2) Corner anchoring: Keep the highest tile in a corner and build around it.\\
Avoid:\\
- Getting high-value tiles stuck in the middle of the board\\
- Creating scattered small values that block potential merges\\
- Making moves that could lead to grid lock\\
Previous Game History:\\
\textcolor{blue}{\{Previous Game History\}}\\
Please analyze the 2048 board and determine the best move.\\
\textcolor{blue}{\{Symbolic Board Features\}}\\
Key considerations:\\
- Look for opportunities to merge similar tiles\\
- Maintain your highest tiles in a corner\\
- Keep space for new tiles to appear\\
- Avoid trapping high-value tiles in the middle\\
IMPORTANT - FORMAT YOUR RESPONSE EXACTLY LIKE THIS:\\
thought: [your analysis here]\\
move: [move]\\
Where [move] must be one of: "up", "down", "left", or "right".\\
Do NOT use \# or any other prefix. Start directly with "thought:" followed by your analysis.

\end{vision-prompt-1}

\begin{vision-prompt-2}

\textbf{\textcolor{blue}{system\_prompt:}}\\
You are an AI agent specialized in 2048 gameplay, your purpose is to analyze board states and suggest optimal moves that maximize your scores.\\

\#\# Your Available Actions\\
For each turn, you must select one command:\\
- up: Shifts the entire grid upward\\
- down: Shifts the entire grid downward\\
- left: Shifts the entire grid leftward\\
- right: Shifts the entire grid rightward\\

When you choose a direction (up, down, left, or right), all tiles shift accordingly. Matching tiles that collide during this shift combine into a single tile representing their sum. After every move, a new tile with a value of either 2 or 4 appears in a random empty cell. The game concludes when no legal moves remain.\\

\textbf{\textcolor{blue}{user\_prompt:}}\\
\#\# 2048 Gameplay Strategies

\#\#\# Principles
The most successful 2048 strategies typically involve:\\
1. Establish your highest-value tile in one corner and build a descending value structure around it.\\
2. Maintain consistent movement patterns that preserve your high-value corner configuration while allowing for regular merges.\\
3. Anticipate how each potential move affects not just the immediate board state but your options 2-3 moves ahead.\\
4. Create opportunities for chain reactions where multiple merges can occur in a single directional move.\\
5. Implement proven arrangements such as:\\
   - Decreasing value snakes that zigzag across the board.\\
   - Corner-anchored structures with decreasing values along the edges.\\
   
\#\#\# Pitfalls to Avoid\\
Certain decisions consistently lead to board deterioration:\\
- Allowing high-value tiles to become isolated in central positions.\\
- Creating scattered low-value tiles that impede potential combinations.\\
- Making moves that reduce overall board fluidity and movement options.\\

\#\# Current Game Context\\
\textcolor{blue}{\{Previous Game History\}}\\

\#\# Board Analysis\\
\textcolor{blue}{\{Symbolic Board Features\}}\\

\#\# Response Protocol\\
**YOUR ANALYSIS MUST STRICTLY ADHERE TO THIS FORMAT:**\\
thought: [Provide your detailed reasoning about the current board state, potential moves, and strategic implications]\\
move: [move]\\

Your move selection must be one of these exact terms: "up", "down", "left", or "right".\\

Begin your response directly with "thought:" followed by your strategic analysis. Do not include any prefixes, headers, or additional formatting.

\end{vision-prompt-2}

\subsection{DSPy Optimized Prompts and Comparison}

\begin{algorithm}[h]
\caption{\textsc{DSPy}: Prompt Optimisation with \textsc{SIMBA}}
\KwIn{%
  Training environments $\mathcal{E}_{\mathrm{train}}$, %
  development environments $\mathcal{E}_{\mathrm{dev}}$, %
  \emph{target} LM set $\mathcal{M}_t$ for performance evaluation,\\
  \emph{optimizer} LM set $\mathcal{M}_o=\{\texttt{o3},\texttt{gemini‑2.5‑pro},\texttt{claude‑3.7‑think},\texttt{deepseek‑R1},\texttt{grok3‑mini}\}$, %
  maximum optimisation steps $k$}
\KwOut{Best prompt module $\mathcal{P}^\star$ (highest mean dev score over all $M_t$)}
\BlankLine

$\mathcal{P}\,\leftarrow\,\texttt{ChainOfThought}(\text{``state}\!\rightarrow\!\text{action''})$
$s_{\mathrm{best}}\leftarrow -\infty$,  $\mathcal{P}^\star\leftarrow\mathcal{P}$\;

\ForEach{$M_o \in \mathcal{M}_o$}{%
  \texttt{dspy.configure(lm=$M_o$)}\;
  
  \textbf{// joint optimisation across \emph{all} target LMs}\\
  $\mathcal{O}\leftarrow\texttt{SIMBA}(\{M_t\},k)$\;
  $\widehat{\mathcal{P}}\leftarrow\mathcal{O}.\texttt{compile}(\mathcal{P},\mathcal{E}_{\mathrm{train}})$\;
  
  \BlankLine
  \textbf{// evaluate average dev score over every $M_t$}\\
  $s_{\mathrm{avg}}\leftarrow 0$\;
  \ForEach{$M_t \in \{M_t\}$}{%
    \texttt{dspy.configure(lm=$M_t$)}\;
    $s_{\mathrm{avg}}\mathrel{+}= \texttt{Evaluate}(\widehat{\mathcal{P}},\mathcal{E}_{\mathrm{dev}})$\;
  }
  $s_{\mathrm{avg}}\leftarrow s_{\mathrm{avg}} / |\{M_t\}|$\;
  
  \If{$s_{\mathrm{avg}} > s_{\mathrm{best}}$}{%
    $s_{\mathrm{best}}\leftarrow s_{\mathrm{avg}}$\;
    $\mathcal{P}^\star\leftarrow\widehat{\mathcal{P}}$\;
  }
}

\Return $\mathcal{P}^\star$
\end{algorithm}

\begin{vision-prompt-dspy1}
\textbf{\textcolor{blue}{system\_prompt:}}\\

You are an AI agent specifically designed to play the game 2048. Your primary objective is to make strategic moves that effectively merge tiles to achieve the highest possible tile value.\\

\textbf{\textcolor{blue}{user\_prompt:}}\\
\#\# Game Overview\\
The game 2048 involves combining identical number tiles on a grid to create tiles with progressively higher values.\\

\#\# Game Mechanics\\
- The game is played on a **4×4 grid**\\
- Each move (up, down, left, right) shifts all tiles in the chosen direction\\
- When two identical tiles collide during a move, they merge into a single tile with twice the value\\
- After each move, a new tile (either 2 or 4) appears randomly in an empty cell\\
- The game concludes when no legal moves remain available\\

\#\# Action Space\\
- **up**: Shifts all tiles toward the top of the grid\\
- **down**: Shifts all tiles toward the bottom of the grid\\
- **left**: Shifts all tiles toward the left side of the grid\\
- **right**: Shifts all tiles toward the right side of the grid\\

\#\# Strategic Principles\\
1. **Corner Anchoring**: Position your highest-value tile in a corner and build around it\\
2. **Structural Stability**: Arrange surrounding tiles in descending order to create a stable formation\\
3. **Maintaining Merge Paths**: Always keep at least one direction available for safe combinations\\
4. **Creating Chain Reactions**: Set up moves that trigger multiple merges in a single action\\
5. **Forward Planning**: Think 2-3 moves ahead to avoid grid-lock and maintain empty spaces\\

\#\# Pitfalls to Avoid\\
- Allowing high-value tiles to drift into central positions\\
- Scattering small-value tiles that obstruct potential merges\\
- Making moves that leave the board with no follow-up merge opportunities\\

\#\# Context Variables\\
\#\#\# Previous Game History\\
\textcolor{blue}{\{Previous Game History\}}\\

\#\#\# Board Features\\
\textcolor{blue}{\{Previous Game History\}}\\

\#\# Response Format\\
Your response must follow this exact two-line format:\\
\begin{verbatim}
```
thought: [your brief analysis of the current board state]\\
move: [up|down|left|right]\\
```
\end{verbatim}
**Important**: Include nothing else beyond these two lines. No additional text, prefixes, symbols, or explanations.\\

\end{vision-prompt-dspy1}

\begin{vision-prompt-dspy2}
\textbf{\textcolor{blue}{system\_prompt:}}

You are an **AI agent** playing **2048**. Your objective is to select moves that merge tiles efficiently and achieve the highest possible tile value.\\

\textbf{\textcolor{blue}{user\_prompt:}}\\
\# 2048 Gaming Guide\\
\#\# Primary Goal\\
Combine like tiles to reach **2048** or higher.\\

---\\

\#\# Game Mechanics\\
- Played on a **4×4 grid**; each move (“up”, “down”, “left”, “right”) shifts every tile.\\
- Identical tiles that collide merge into one tile with **double the value**.\\
- After each move, a new tile (**2** or **4**) appears randomly in an empty cell.\\
- The game ends when **no legal moves** remain.\\

---\\

\#\# Action Space\\
- **up**: Shift all tiles upward.\\
- **down**: Shift all tiles downward.\\
- **left**: Shift all tiles to the left.\\
- **right**: Shift all tiles to the right.\\

---\\

\#\# Key Strategies\\
1. **Corner anchor** - Park your highest tile in one corner and build around it.\\
2. **Stable structure** - Arrange surrounding tiles in descending order to protect the corner.\\
3. **Clear merging path** - Keep at least one direction available for safe combinations.\\
4. **Chain reactions** - Set up moves that trigger multiple merges in one swipe.\\
5. **Look ahead** - Plan 2–3 moves in advance to avoid grid-lock. Preserve empty spaces for new tiles.\\

---\\

\#\# Avoid\\
- Letting high-value tiles drift into the center.\\
- Scattering small tiles that block merges.\\
- Moves that leave the board with no follow-up merges.\\

---\\

\#\# Previous Game History\\
\textcolor{blue}{\{Previous Game History\}}\\

\#\# Board Features\\
\textcolor{blue}{\{Symbolic Board Features\}}\\

**Response format (use exactly two lines):**\\
thought: your brief analysis of the current board\\
move: up | down | left | right\\

Include nothing else—no prefixes, symbols, or extra text.

\end{vision-prompt-dspy2}

\subsection{Performance Comparison}
In our agent, we follow the SIMBA optimizer implementation in DSPy with performance metrics defined in Section~\ref{sec:game_metrics} to optimize the prompt with five optimizer models: Claude-3-7-sonnet, Gemini-2.5-Pro-Preview, o3, Deepseek-r1, Grok-3-Mini-Beta, to search the best performing prompt yielding highest average reward cross all target models in 20 optimization steps ($k=20$).

\begin{table}[h]
    \centering
    \scriptsize
    \caption{Model performance across various prompt types in Game 2048 with harness, where $\Delta_e$ and $\Delta_p$ stand for performance difference between empirically deisgn prompt pairs and DSPy optimzied prompt pairs. P1 and P2 denotes to different prompt templates.}
     \resizebox{\linewidth}{!}{%
    \begin{tabular}{l|ccc|ccc}
        \toprule
        \textbf{Model} & \textbf{Empirical P1} & \textbf{Empirical P2} & $ {\lvert}\Delta_e {\rvert} \left(\downarrow \right)$ & \textbf{DSPy P1} & \textbf{DSPy P2} &${\rvert}\Delta_p {\lvert} \left(\downarrow \right)$\\
        \midrule
        gemini-2.5-flash-preview-04-17       &  1697.3{\scriptsize$\pm$548}                   &        1478.7{\scriptsize$\pm$440}   &  218.6 &      1746.0{\scriptsize$\pm$518}   &         1601.3{\scriptsize$\pm$174}       &   \textbf{144.7}  \\
        claude-3-5-sonnet-20241022        &            2624.0{\scriptsize$\pm$466}             &            2235.3{\scriptsize$\pm$862}   &  388.7  &      2786.0{\scriptsize$\pm$290}          &    2928.0{\scriptsize$\pm$318}   &       \textbf{142.0}       \\
        o4-mini-2025-04-16        &           4432.0{\scriptsize$\pm$1096}              &               3680.0{\scriptsize$\pm$963}   &     752.0
        &      3851.3{\scriptsize$\pm$864.4}    &   4320.0{\scriptsize$\pm$700}  &       \textbf{468.7}         \\
        \bottomrule
    \end{tabular}%
    }
    \vspace{0.5em}
    \label{tab:prompt_optimization}
\end{table}


Among three target models: Gemini-2.5-Flash-Preview, Claude-3-5-Sonnet, o4-mini, experiments results show evidence that our prompt optimization pipeline can reduce performance discrepancy between two candidate prompts by 33.8\% to 63.5\% on the three models across 3 runs. Details are reported in Table~\ref{tab:prompt_optimization}.

\end{document}